\documentclass[journal]{IEEEtran}
\let\labelindent\relax
\usepackage{xcolor,soul,framed} %,caption
\colorlet{shadecolor}{yellow}
\usepackage[pdftex]{graphicx}
\graphicspath{{../pdf/}{../jpeg/}}
\DeclareGraphicsExtensions{.pdf,.jpeg,.png}

\usepackage[cmex10]{amsmath}
\usepackage{array}
\usepackage{mdwmath}
\usepackage{mdwtab}
\usepackage{eqparbox}
\usepackage{url}
\usepackage{enumitem}
\usepackage{multirow}
\usepackage{amsmath} 
\usepackage{amssymb}
\usepackage{mathrsfs}
\usepackage{pifont}
\usepackage{bbding}
\usepackage{threeparttable}
\usepackage{booktabs}

\usepackage{graphicx}

\usepackage[caption=false,font=footnotesize,labelfont=rm,textfont=rm]{subfig}
\usepackage{stfloats}
\usepackage{cite}
\begin{document}
\bstctlcite{IEEEexample:BSTcontrol}
    % \title{Hyperspectral Adapter for Hyperspectral Object Tracking}
    \title{Hyperspectral Adapter for Object Tracking based on Hyperspectral Video}
\author{
      Long Gao\textsuperscript{a*},
      Yunhe Zhang\textsuperscript{a},
      Langkun Chen\textsuperscript{a},
      Yan Jiang\textsuperscript{b}, 
      Weiying Xie\textsuperscript{a}, 
      Yunsong Li\textsuperscript{a},
      
\thanks{\textsuperscript{a} State Key Laboratory of Integrated Service Networks, School of Telecommunications Engineering, Xidian University, No.2, South Taibai Street, Hi-Tech Development Zone, Xi’an, China, 710071.} 
\thanks{\textsuperscript{b} The Department of Electronic and Electrical Engineering, the University of Sheffield, Sheffield, UK, S10 2TN.}
\thanks{\textsuperscript{*} Corresponding author: \emph{Long Gao}}
% \thanks{\textsuperscript{*} Corresponding author: \emph{Weiying Xie}}
}
% \markboth{IEEE TRANSACTIONS ON GEOSCIENCE AND REMOTE SENSING, VOL. X, NO. X, X 2023
% }{Roberg \MakeLowercase{\textit{et al.}}: High-Efficiency Diode and Transistor Rectifiers}

\markboth{IEEE TRANSACTIONS ON CIRCUITS AND SYSTEMS FOR VIDEO TECHNOLOGY
}{Roberg \MakeLowercase{\textit{et al.}}: High-Efficiency Diode and Transistor Rectifiers}

% \markboth{IEEE TRANSACTIONS ON GEOSCIENCE AND REMOTE SENSING, VOL. X, NO. X, X 2025
% }{Roberg \MakeLowercase{\textit{et al.}}: High-Efficiency Diode and Transistor Rectifiers}

% \markboth{IEEE TRANSACTIONS ON NEURAL NETWORKS AND LEARNING SYSTEMS
% }{Roberg \MakeLowercase{\textit{et al.}}: High-Efficiency Diode and Transistor Rectifiers}
% IEEE TRANSACTIONS ON NEURAL NETWORKS AND LEARNING SYSTEMS

% ====================================================================
\maketitle

% === ABSTRACT ====================================================================
% =================================================================================
\begin{abstract}
%\boldmath
Object tracking based on hyperspectral video attracts increasing attention to the rich material and motion information in the hyperspectral videos. The prevailing hyperspectral methods adapt pretrained RGB-based object tracking networks for hyperspectral tasks by fine-tuning the entire network on hyperspectral datasets, which achieves impressive results in challenging scenarios.
However, the performance of hyperspectral trackers is limited by the loss of spectral information during the transformation, and fine-tuning the entire pretrained network is inefficient for practical applications.
To address the issues, a new hyperspectral object tracking method, hyperspectral adapter for tracking (HyA-T), is proposed in this work. 
The hyperspectral adapter for the self-attention (HAS) and the hyperspectral adapter for the multilayer perceptron (HAM) are proposed to generate the adaption information and to transfer the multi-head self-attention (MSA) module and the multilayer perceptron (MLP) in pretrained network for the hyperspectral object tracking task by augmenting the adaption information into the calculation of the MSA and MLP. 
Additionally, the hyperspectral enhancement of input (HEI) is proposed to augment the original spectral information into the input of the tracking network. The proposed methods extract spectral information directly from the hyperspectral images, which prevent the loss of the spectral information.
Moreover, only the parameters in the proposed methods are fine-tuned, which is more efficient than the existing methods. Extensive experiments were conducted on four datasets with various spectral bands, verifing the effectiveness of the proposed methods. The HyA-T achieves state-of-the-art performance on all the datasets.
The source codes and model are available at https://github.com/lgao001/HyA-T.

\end{abstract}

% === KEYWORDS ====================================================================
% =================================================================================

% \fi
\IEEEpeerreviewmaketitle

% ====================================================================
% ====================================================================
% ====================================================================

% === I. INTRODUCTION =============================================================
% =================================================================================
\section{Introduction}

\IEEEPARstart{R}{ecently,} tracking object with hyperspectral (HS) videos has drawn increasing interest in the field of visual object tracking since the implicit material information in the HS images benefits the accurate tracking in challenge scenarios \cite{1,2}. While significant efforts have been devoted to object tracking using RGB videos, and considerable progress has been made, the performance of these trackers remains unsatisfactory in difficult situations \cite{3,4,5}, such as background clutter, rapid deformation and occlusion. Different from the RGB images, HS images consist of more detailed spectral information, which can be utilized to identifying the material information \cite{6}. Therefore, increasing attention has been drawn in the research on the HS object tracking. 

The leading framework for the HS object tracking is based on deep learning. However, directly constructing the HS object tracking system and training the network using HS datasets achieved unsatisfactory results since the quantity of the HS training data was insufficient to train a robust tracking network \cite{7}. Consequently, the existing methods transferred the pretrained RGB-based object tracking networks into the HS object tracking task to inherit the powerful discriminative ability, which is learned from a large quantity of RGB data \cite{8,9,10}. Specifically, the HS images were converted into either a single three-channel false-color image through band selection or multiple three-channel false-color images using band regrouping to adapt to the transferred tracking networks. The pretrained tracking networks were fine-tuned using the converted images. \cite{11,12,13,14}. By processing the converted HS images with the fine-tuned network, the object tracking with HS videos was realized. Comparing to the pretrained RGB-based object tracking networks, the HS image-based tracking methods achieved better performance in the same challenging scenario, which confirms the effectiveness of the pretraining-and-fine-tuning methods \cite{15,16,17}. Moreover, tracking with multiple false-color images demonstrates improved performance in existing HS object tracking methods, indicating that more spectral information utilization benefits the performance of the trackers \cite{18,19}. Recently, the transformer-based object tracking networks were transferred into the multiple false-color images based HS object tracking methods since their impressive performance in the RGB-based object tracking task. State-of-the-art performance was achieved by the transformer-based HS object tracking method due to the long-range dependency capturing ability of the transformer \cite{20,21,22}.

% =======
% FIG. 01
% =======
% \begin{figure}
%   \begin{center}
%   \includegraphics[width=3.5in]{pdf-prompt/figure1-2.pdf}\\
%  \caption{Comparison of different types of methods for hyperspectral tracking. BS represents the band selection module. BR represents the band regrouping module. Frozen stands for that the 
 % parameters in the modules are initialized with pretrained parameters and frozen during the training procedure. 
%  parameters within the modules are set using pretrained values and remain unchanged during training process.
%  Fine-tune refers to fine-tuning the parameters.}\label{Fig1}
%   \end{center}
% \end{figure}

Despite the merits, the performance improvement of the HS object tracking methods is limited by the loss of spectral information. Specifically, converting the HS images, which consist of 16 or more spectral bands, into one false-color image resulted in extensive spectral information loss, and worse performance than the trackers applying multiple false-color images \cite{11,12}. Moreover, although more spectral information was utilized, the bands regrouping methods in the trackers based on multiple false-color images caused the loss of the interactive information between the bands \cite{15,20}. Moreover, the fine-tuning methods consumed a large amount of memory, which was unfriendly for application. The fine-tuning methods in the existing HS methods updated the parameters in the pretrained network and the additional modules designed for the HS object tracking task. For the trackers based on multiple false-color images, multiple sets of parameters were trained for the false-color images, which depended on the number of the false-color images, e.g., 8 sets of parameters for the HS images with 25 channels \cite{20,21,22}.

In recent years, parameter-efficient fine-tuning (PEFT) has gained significant attention in computer vision tasks \cite{78} due to the effectiveness of the transferring a based model into a downstream task. Specifically, in the visual object tracking task, the methods in \cite{79} and \cite{80} transferred the base model, RGB object tracking network, into the downstream tasks, object tracking with satellite videos and under-water videos. However, the PEFT methods still struggle to fully leverage the rich spectral information inherent in HS data due to the high dimension nature of the HS images.

To address the limitations of developing robust HS object tracking methods, a novel and accurate transformer-based HS object tracking method, hyperspectral-adapter-based tracking (HyA-T), is proposed. Firstly, the hyperspectral adapter for self-attention (HAS) is proposed to adapt the pretrained tracking network for the HS object tracking task by augmenting the HS information into the features, which are extracted from the encoder layers in the pretrained object tracking network. HyA-T is constructed using a transformer-based object tracking network that utilizes a ViT to extract features \cite{23}. The HAS generates the adaption information from the HS images, and augments the adaption information into the queries and values in the self-attention learning of the ViT. Different from the existing methods, which learns the spectral information from the converted HS images, the HAS learns the spectral information directly from the HS images, avoiding the loss of spectral information caused by the transformation. Secondly, the hyperspectral adapter of multilayer perceptron (HAM) is introduced to adapt the multilayer perceptron (MLP) in the ViT for the HS object tracking task. The MLP is a crucial part of the ViT, and consists of similar number of parameters to the self-attention learning module. To avoid directly fine-tuning the MLP, the HAM implements two lightweight structures, which are parallel and sequentially to the MLP respectively. By training the HAM with the HS training data, the HAM adapts the output of the MLP to the HS object tracking task. Thirdly, the hyperspectral enhancement of the input (HEI) is proposed to generate the input of the tracking network, which contains the intact spectral information in the HS images. In the HEI, the HS image is converted into a three-channel false-color image with the CIE color matching functions (CMFs) \cite{1} since the pretrained object tracking network takes the three-channel image as input. Then, the HEI fuses the HS image and the false-color image to augment the spectral information into the input of the tracking network, which reduces the negative influence of the loss of the spectral information. Moreover, the HAS, HAM and HEI are constructed with the lightweight structure, and only the parameters in the three modules are fine-tuned on HS dataset. In this way, the quantity of the parameters needed for fine-tuning is significantly reduced.

Extensive experiments are conducted on four HS object tracking tasks, i.e., HOTC \cite{1}, HOTC2024-VIS \cite{24}, HOTC2024-NIR \cite{24}, HOTC2024-RedNIR \cite{24}. The experimental results verify the effectiveness of the proposed methods. The HyA-T achieves the state-of-the-art performance in all the four datasets. Moreover, only 5$\%$ on parameters in the network are needed to be trained on the HS training datasets. The contributions of this work can be summarized as follows:

\begin{itemize} [leftmargin=1.3em]

\item A novel transferring method, the HAS is proposed to build an accurate HS object tracking method. Comparing with the existing HS object tracking methods, HAS preserves both spectral and interactive information without any loss.

\item An adapter for the MLP in encoder layers of the pretrained network, a hyperspectral adapter for MLP (HAM), is proposed. By implementing the adapter parallel and sequentially to the MLP, the HAM adapts the MLP to the HS object tracking task in parameter-efficient manner.

\item The HEI is introduced to generate the input containing spectral information for the transferred object tracking network. HEI enhances the unaltered spectral information in HS images, incorporating it into the network's input, which decreases the loss of spectral information.

\item Based on the proposed methods, a new HS object tracking method, HyA-T, is proposed in this work. The proposed method requires training for less than 5$\%$ of the parameters in the HS object tracking network, significantly fewer than the parameters required by existing methods that use multiple false-color images. Comprehensive experiments are conducted, and state-of-the-art performance is achieved by HyA-T in multiple HS object tracking datasets, i.e., 0.705 and 0.776 in AUC on HOTC and HOTC2024-NIR, respectively.

\end{itemize}

% === II. Harmonically-Terminated Power Rectifier Analysis ========================
% =================================================================================
\section{Related Work}

This section first reviews RGB object tracking methods, which serve as the foundation for HS tracking approaches. Subsequently, existing HS object tracking algorithms are summarized. Finally, parameter-efficient fine-tuning (PEFT) techniques, i.e., LoRA and prompt-based methods, are introduced as approaches closely related to our study.

\subsection {RGB Object Tracking Methods}

In recent years, significant progress has been made in RGB object tracking methods. Many of these advancements have been achieved through approaches based on the Siamese network framework, which formulates the tracking task as a matching problem between a template and search regions \cite{72}. Early works employed convolutional neural network (CNN) and correlation operations to track the target \cite{25,26,27,28}. Improvements, i.e., region proposal network (RPN) \cite{29,30}, anchor-free methods \cite{31,32,33}, and attention mechanisms \cite{34,35,36,37,73} were utilized in the tracking networks to achieve better performance. Recently, transformer was introduced in the object tracking networks for the long-range dependency information capturing ability of the transformer, and better performance was achieved \cite{38,39,76}. TransT and STARK implemented the transformer to enhance the features extracted with the CNN, and modeled the correlation operations with transformer \cite{40,41}. Then, Mixformer, SwinTrack and SFTransT further employed the transformer-based networks as feature extractors, leveraging the hierarchical self-attention structure to enhance multi-scale feature learning \cite{3,10,74}. The tracking networks mentioned above conducted the feature extraction and the correlation modeling separately, which had limited target-background discriminability. Consequently, OSTrack and SeqTrack unified the feature extraction and correlation modeling, and proposed concise tracking networks \cite{42,43}. Based on that, ARTrack, HIPTrack, AQATrack, and TaTrack further exploited the template updating strategies to obtain more robust tracking results \cite{44,45,46,75}. Despite the significant progress in the RGB object tracking methods, some challenging scenarios are still hard for the tracking methods, and increasing interests are drawn to perform the object tracking with HS images. 

\subsection {Hyperspectral Object Tracking Methods}

HS images provide significant advantages for object tracking due to their rich spectral information, which benefits for the robust object tracking in complex scenarios, particularly when visual cues in RGB images are insufficient \cite{1}. However, feature extraction with high dimensional images and limited training data complicated the development of effective HS tracking models \cite{47}. To address these challenges, some methods relied on handcrafted features to extract spectral-spatial information \cite{1,48,49}. Xiong et. al. integrated spectral unmixing techniques with spectral-spatial feature extraction methods and coupling them with a correlation filter-based tracker \cite{1}. Although these methods demonstrated the potential of HS tracking, their reliance on handcrafted low-level features limited their effectiveness in handling complex scenarios. To address these challenges, RGB object tracking methods have been transferred for HS applications by fine-tuning pretrained models on HS datasets. To adapt the RGB object tracking networks, SSDT-Net \cite{50} and SiamHT \cite{12} transformed HS images into one three-channel false-color image to enable the use of pretrained RGB tracking networks, achieving notable improvements in processing speed. However, the reduced spectral detail in this approach constrained its tracking performance. Consequently, spectral band regrouping organized the bands into multiple false-color images based on their relevance to target discrimination, preserving more spectral information and enhancing tracking performance \cite{18,19}. Some subsequent methods have optimized band selection and integrated feature fusion to significantly improve efficiency and accuracy \cite{13,15,20}. These approaches improved the performance of the HS object tracking methods. However, the conversion of the HS images leads to a loss of spectral information, and the limited number of the HS training data still restricted the further improvement of the HS object tacking methods. 

\subsection {Parameter-Efficient Fine-Tuning in Visual Tracking}

Parameter-efficient fine-tuning (PEFT), which adapts pretrained models to downstream tasks by freezing most parameters and introducing lightweight trainable modules, was initially developed for natural language processing (NLP) tasks, and had become a widely adopted method in large language models (LLMs) \cite{51,52}. Among these, PEFT techniques such as Prompt Tuning and Low-Rank Adaptation (LoRA) have also achieved significant success in computer vision tasks \cite{53,54}. ProTrack \cite{55} and ViPT \cite{56} incorporated learnable parameters into prompts, enabling better feature alignment and delivering significant improvements in multi-modal object tracking task. Moreover, OneTracker introduced prompt-based methods for more combination of modalities, broadening the adaptability of tracking methodologies to diverse modalities \cite{57}. Apart from that, PromptVT employed prompt-based mechanisms to dynamically update templates throughout the tracking process \cite{77}. Beyond Prompt Tuning, LoRA enables efficient transfer of pretrained tracking models to other tasks by introducing low-rank parameter updates to specific layers, which preserves core pretrained capabilities and adapts to specialized scenarios with minimal additional computational cost \cite{52}. The BAT implemented the adapters to transfer the RGB tracking network to the multi-modal object tracking task \cite{58}. Different from the existing prompt-based and the LoRA-based methods in multi-modal object tracking task, our research explores the potential of adapting pretrained RGB object tracking models to HS tracking task by leveraging the modified version of adapter, which generates the adaption information from the HS images.

% === III. Schottky-Diode Class-C Rectifier =======================================
% =================================================================================
\section{Proposed Method}

This section outlines the detailed framework of the proposed HyA-T method. First, there is a brief introduction of the overall framework of HyA-T. Then the detailed work of its core elements: the hyperspectral adapter for self-attention (HAS), the hyperspectral adapter for MLP (HAM), and the hyperspectral enhancement of input (HEI) is discussed. Finally, the training and inference strategies of HyA-T are elaborated to demonstrate the method's practical implementation.

% =======
% FIG. 01
% =======
\begin{figure*}
  \begin{center}
  \includegraphics[width=7.25in]{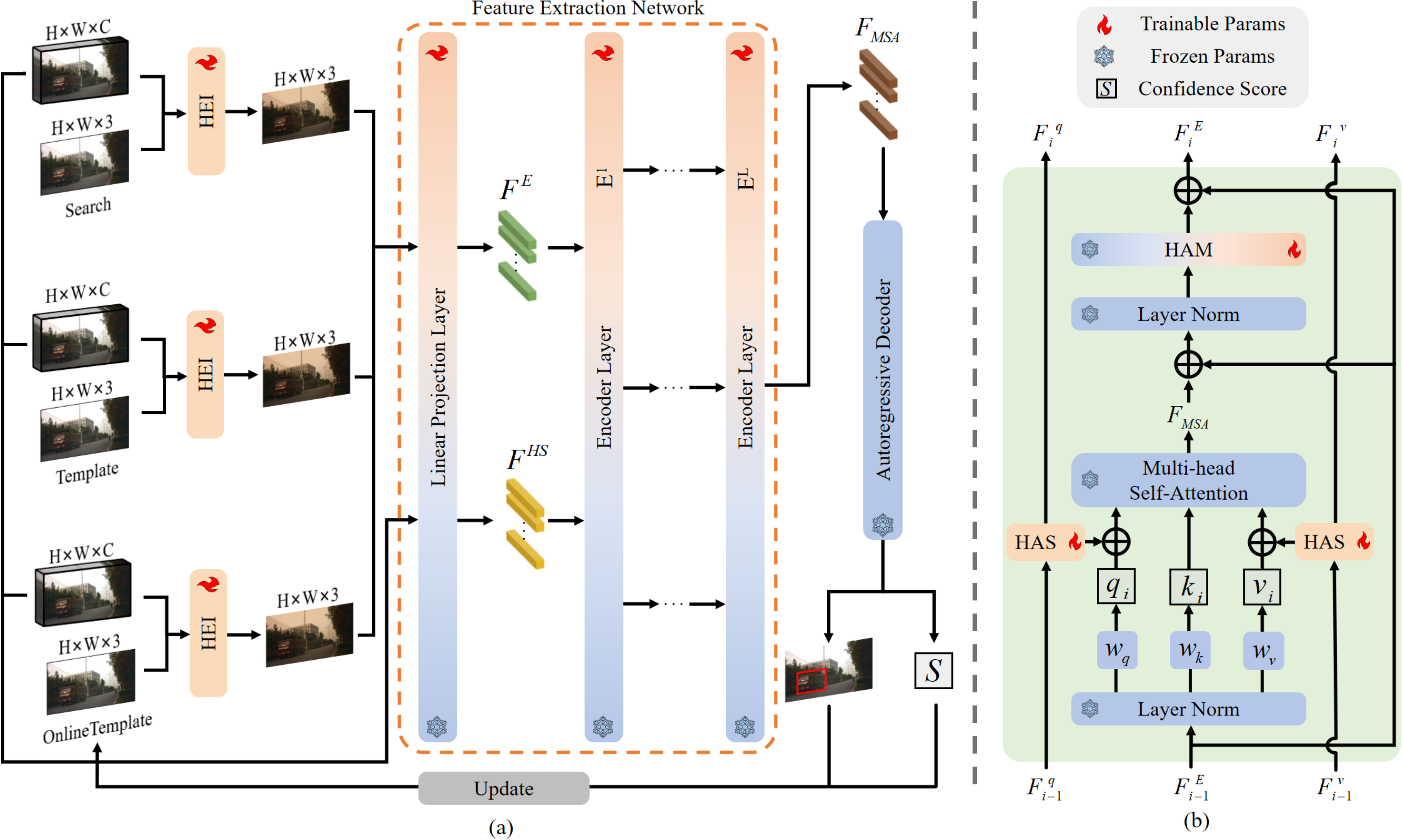}\\
  % \vspace{1mm}
  % \begin{flushleft}
  %     { \hspace{63mm} (a) \hspace{84mm} (b) }
  %   \end{flushleft}
 \caption{(a) The pipeline of HyA-T. The overall structure is composed of HEI, a feature extraction network, and an autoregressive decoder. (b) The internal structure of a transformer block in the encoder, with the incorporation of HAS and HAM.}\label{Fig1}
  \end{center}
\end{figure*}

\subsection {Overall Framework}

HyA-T contains three stages, which are HEI for the input augmentation, HAS and HAM for the feature extraction and the autoregressive decoder for the estimation of the bounding box of the target. Fig.~\ref{Fig1} presents the pipeline of the HyA-T, and the description of the modules are as below.

\subsubsection{\textbf{HEI}}

As the inputs of HyA-T, three pairs of images, i.e., the template pair, the online template pair, and the search pair are first processed by the HEI. The template pair is extracted from the first frame of the HS video, capturing the initial appearance of the target. The online template pair is used to dynamically update the target’s representation during the tracking. The search pair represents the regions containing the target in current frame. In each image pair, there are two images, i.e., the HS image and the false-color image, e.g., ${I_{S}^{HS}} \in \mathbb{R} {^{H \times W \times C}}$ and ${I_{S}^{FRGB}} \in \mathbb{R} {^{H \times W \times 3}}$ in the search pair. The false-color image is obtained by the CMFs. Here, $H$, $W$, and $C$ signify the height, width, and the number of channels, respectively. The application of HEI for the search pair is described as follows:
\begin{equation}\label{Eq1}
    I_S^{E} = HEI(I_S^{HS},I_S^{FRGB})
\end{equation}
where ${I_{S}^{E}} \in \mathbb{R} {^{H \times W \times 3}}$ denotes the enhanced false-color image of the search pair and $HEI(\cdot )$ represents the application of the HEI module. Additionally, the template pair and the online template pair also undergo same process, yielding their corresponding outputs ${I_{T}^{E}} \in \mathbb{R} {^{H \times W \times 3}}$ and ${I_{O}^{E}} \in \mathbb{R} {^{H \times W \times 3}}$.
% where ${I_{S}^{E}} \in \mathbb{R} {^{H \times W \times 3}}$ denotes the enhanced false-color image of the search pair and $HEI(\cdot )$ represents the application of the HEI module. Additionally, the template pair and the online template pair also undergo processing by the HEI module, yielding their corresponding outputs ${I_{T}^{E}} \in \mathbb{R} {^{H \times W \times 3}}$ and ${I_{O}^{E}} \in \mathbb{R} {^{H \times W \times 3}}$.

\subsubsection{\textbf{Feature Extraction Network}}

As shown in Fig.~\ref{Fig1}, the feature extraction network comprises one linear projection layer followed by multiple encoder layers, with each of the encoder layer further incorporating the HAS and HAM modules.

The primary task of the linear projection layer is to transform the images from three sets, i.e., the template set, the online template set and the search set, into patch features. Taking the search set as an example, the HS image ${I_{S}^{HS}}$ and the corresponding enhanced false-color image ${I_{S}^{E}}$ are passed through the linear projection layer, where they are processed separately through distinct patch embedding branches and flattened into two specific patch vectors, ${F_{S}^{HS}} \in \mathbb{R} {^{N \times D}}$ and ${F_{S}^{E}} \in \mathbb{R} {^{N \times D}}$. Here, $N$ denotes the number of tokens extracted from a single image, and $D$ represents the channels of each token. In parallel, the template set and the online template set are also processed by this layer, resulting in their respective patch vectors ${F_{T}^{HS}},{F_{T}^{E}} \in \mathbb{R} {^{N \times D}}$ and ${F_{O}^{HS}},{F_{O}^{E}} \in \mathbb{R} {^{N \times D}}$. Then these vectors from the three sets are concatenated to form the HS vectors, ${F^{HS}} \in \mathbb{R} {^{3N \times D}}$ and the enhanced false-color vector, ${F^{E}} \in \mathbb{R} {^{3N \times D}}$.

The encoder layers are employed to simultaneously extract features from the template and search regions. Within the encoder layer, ${F^{E}}$ is processed through a multi-head self-attention (MSA) module. Concurrently, ${F^{HS}}$ is iteratively calculated through the HAS. Each output of HAS is then integrated into the attention mechanism of the MSA module. This approach can be represented as:
% \begin{equation}\label{Eq2}
% \begin{aligned}
% &Q,K,V = Linear(F^{E}) \\
% F_{MSA} = MSA(&Q+HAS(F^{HS}),K,V+HAS(F^{HS}))
% \end{aligned}
% \end{equation}
% \begin{small}
\begin{equation}\label{Eq2}
\begin{aligned}
    Q&, K, V = Linear(F^{E}) \\
    F_{MSA} &= MSA( Q + HAS(F^{HS}) , \\
    &K, V + HAS(F^{HS}))
\end{aligned}
\end{equation}
% \end{small}
where $Linear(\cdot )$ represents the linear transformation layer in the attention mechanism. $Q$, $K$, and $V$ are the query, key, and value matrices. ${F_{MSA}} \in \mathbb{R} {^{3N \times D}}$
 refers to the final output of the MSA module. Subsequently, the output feature map, ${F_{MSA}}$  is fed into the modified multilayer perceptron (MLP) module, HAM, and output the final feature, ${F} \in \mathbb{R} {^{3N \times D}}$. The operation can be described as follows.
\begin{equation}\label{Eq3}
F = HAM(F_{MSA})
\end{equation}
where $HAM(\cdot )$ is the operations of HAM module.

\subsubsection{\textbf{Autoregressive Decoder}}

The autoregressive decoder operates on the search region's feature ${F_{S}} \in \mathbb{R} {^{N \times D}}$, which is obtained from $F$ and embeds information about the search image, resulting in the estimation of the bounding box of the target, $B$. In addition to the bounding boxes, the autoregressive decoder outputs confidence scores, $S$, to drive the adaptive online template update mechanism. The process can be expressed as:
\begin{equation}\label{Eq4}
\begin{aligned}
&F_{S} = F[0:N-1] \\
&B,S = f_{decoder}(F_{S}) \\
T_{next} &= Temp\_upd(T_{now},S)
\end{aligned}
\end{equation}
where ${f_{decoder}(\cdot )}$ represents the transformation applied by the autoregressive decoder module. ${Temp\_upd(\cdot )}$ refers to the function of template update mechanism. ${T_{now}}$ and ${T_{next}}$ represent the online template at the current frame and the next one. More details on the autoregressive decoder can be found in \cite{43}. 

% =======
% FIG. 02
% =======
\begin{figure}
  \begin{center}
  \includegraphics[width=1.5in]{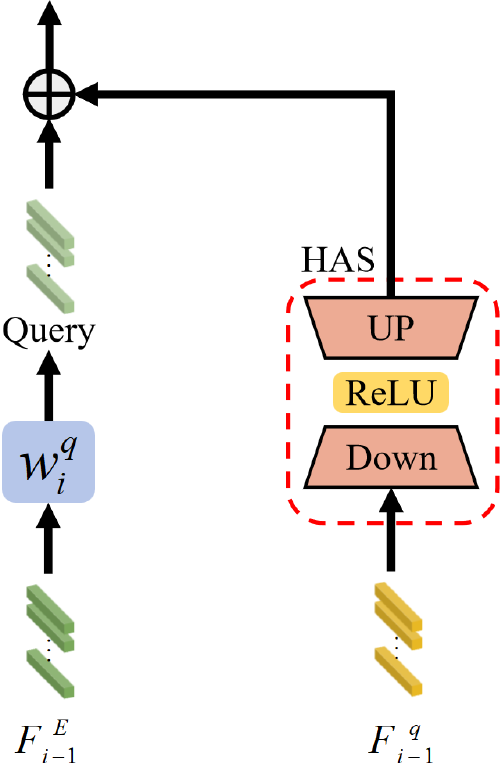}\\
 \caption{The structure of HAS applied to the query branch. The blue component represents the modules kept frozen during training, while the red components indicate the modules that are updated during the training process.}\label{Fig2}
  \end{center}
\end{figure}

\subsection {Hyperspectral Adapter for Self-attention (HAS)}

HAS is proposed to adapt the features in the encoder layers for the HS object tracking task. Specifically, the HAS learns the adaption information from the original HS images, and augments the adaption information into the features extracted from the false-color images, which contains limited spectral information. Moreover, instead of directly fine-tuning the parameters in the MSA, the parameters in the HAS is trained with the HS data, which is parameter-efficient to adapt the MSA for the HS object tracking task.

In the MSA, the query, key, and value vectors for MSA calculation are computed as follows:
\begin{equation}\label{Eq5}
\begin{aligned}
q_{i} &= F_{i}^{E}w_{i}^{q} \\
k_{i} &= F_{i}^{E}w_{i}^{k} \\
v_{i} &= F_{i}^{E}w_{i}^{v}
\end{aligned}
\end{equation}
where ${F_{i}^{E}} \in \mathbb{R} {^{3N \times D}}$ denotes the input of the $i$-th encoder layer, with ${F_{1}^{E}=F^{E}}$. ${q_{i}},{k_{i}},{v_{i}} \in \mathbb{R} {^{12 \times 3N \times (D/12)}}$ are the representations of those vectors in the 12-head MSA, and ${w_{i}^{q}},{w_{i}^{k}},{w_{i}^{v}}$ represent the respective projection weight matrix in the $i$-th encoder layer.

As Fig.~\ref{Fig1}(b) shows, two instances of HAS are incorporated within the MSA of each encoder layer. One is applied to enhance the query and the other is utilized to enhance the value. Fig.~\ref{Fig2} illustrates the internal structure of a HAS module applied to the query branch. It is designed to extract the spectral information through a low-rank transformation module iteratively in each encoder layer. The module begins with a linear projection to effectively condensing the $i$-th HAS module’s input, ${F_{i-1}^{q}} \in \mathbb{R} {^{3N \times D}}$, into a compact token vector with ${F_{0}^{q}=F^{HS}}$. Subsequently, the transformed representation passes through a non-linear activation function to enhance its complexity and adaptability. Finally, another linear projection layer is applied to restore the feature representation to its original dimension scale. In all, the process in the query branch can be formally articulated as:
\begin{equation}\label{Eq6}
\begin{aligned}
% &F_{0}^{q} = F^{HS} \\
F_{i}^{q} = L_{up}&(ReLU(L_{down}(F_{i-1}^{q}))) \\
&{q_{i}^{’}} = q_{i} + F_{i}^{q}
\end{aligned}
\end{equation}
where ${L_{down}(\cdot )}$ and ${L_{up}(\cdot )}$ represent the two linear projection operations with rank 16, ${ReLU(\cdot )}$ is the active function, and ${q_{i}^{’}} \in \mathbb{R} {^{12 \times 3N \times (D/12)}}$ denotes the new query vector with spectral feature integrated. At the same time, a similar process occurs in the value branch with ${F_{0}^{v}=F^{HS}}$, resulting in ${v_{i}^{’}} \in \mathbb{R} {^{12 \times 3N \times (D/12)}}$. Then MSA module calculates the self-attention with ${q_{i}^{’}}$,${k_{i}}$ and ${v_{i}^{’}}$, ultimately producing the final feature ${F_{MSA}}$. The MSA computation is given by:
\begin{equation}\label{Eq7}
\begin{aligned}
F_{i}^{E} &= MSA({q_{i}^{’},k_{i},v_{i}^{’}}) \\
&F_{MSA} = F_{L}^{E}
\end{aligned}
\end{equation}
where $L$ is the total number of encoder layers and $L=12$ in HyA-T.

\subsection {Hyperspectral Adapter for MLP (HAM)}

An encoder layer in the transformer model, e.g., ViT, consists of a MSA module and a MLP module. The MSA is transferred for the HS object tracking task by the HAS. Comparing to the MSA, the MLP is more effective for the high-frequent information \cite{59}. Consequently, the HAM is proposed to adapt the MLP to the HS object tracking.

\begin{figure}
  \begin{center}
  \includegraphics[width=2in]{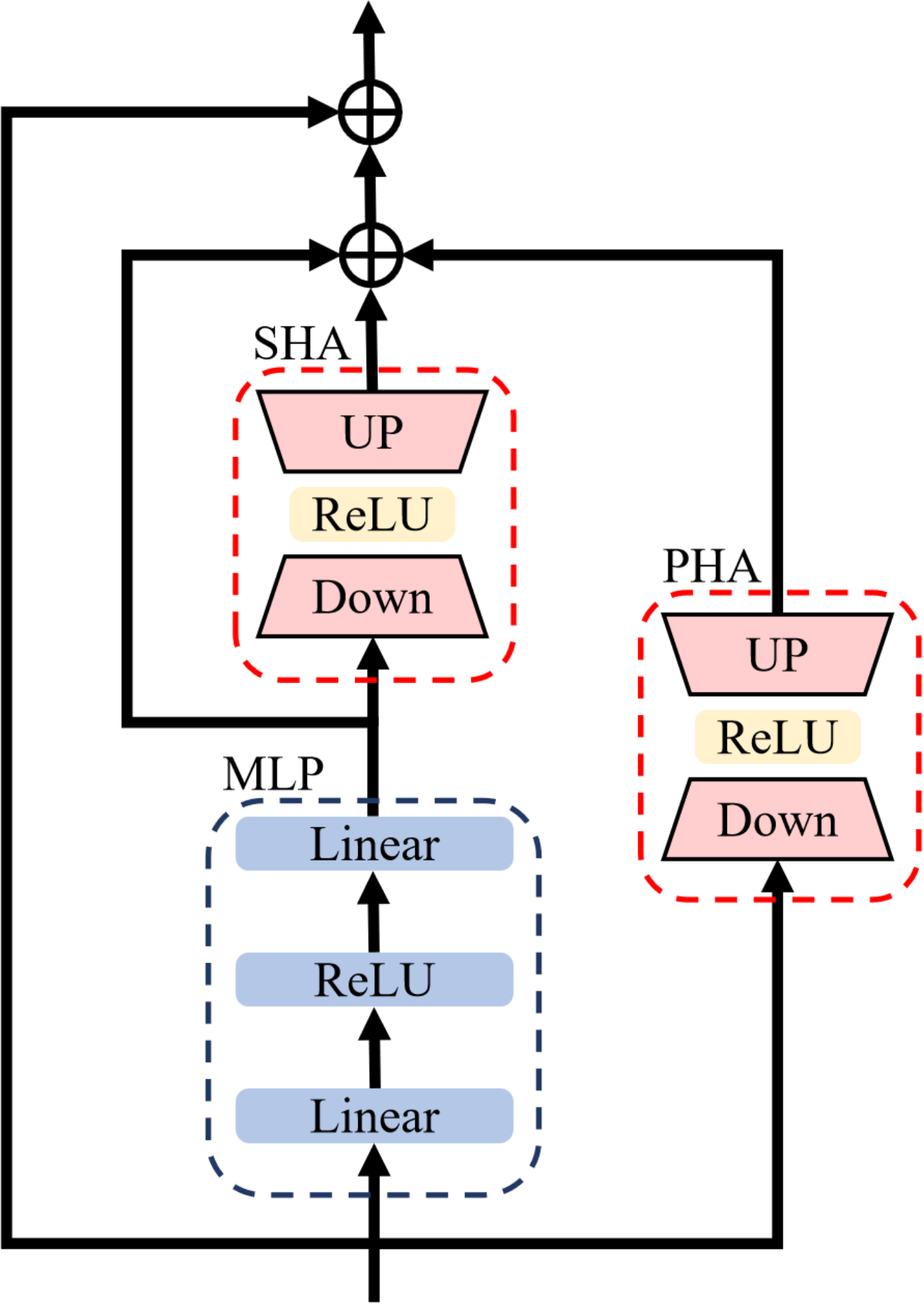}\\
 \caption{The schematic diagram illustrates the operation of HAM. The blue parts stand for the modules that are frozen during the training, and the red parts are updated during the training. The SHA represents the adapter that implemented sequentially to the MLP. The PHA is the adapter that implemented parallelly to the MLP.}\label{Fig3}
  \end{center}
\end{figure}

The structure of the HAM is illustrated in Fig.~\ref{Fig3}. There are two types of HS adapters in the HAM, i.e., sequential hyperspectral adapter (SHA) and parallel hyperspectral adapter (PHA). The input of the HAM is the output of the corresponding MSA, ${F_{MSA}} \in \mathbb{R} {^{3N \times D}}$. ${F_{MSA}}$ is processed with MLP and PHA as follows.
\begin{equation}\label{Eq8}
\begin{aligned}
F_{MLP} = MLP(F_{MSA})
\end{aligned}
\end{equation}
\begin{equation}\label{Eq9}
\begin{aligned}
F_{PHA} = UP_{PHA}(ReLU(Down_{PHA}(F_{MSA})))
\end{aligned}
\end{equation}
where ${F_{MLP}} \in \mathbb{R} {^{3N \times D}}$ and ${F_{PHA}} \in \mathbb{R} {^{3N \times D}}$  denote the output of original MLP module and the PHA, ${MLP(\cdot )}$ represents the regular MLP operation which includes two linear layers and an activation layer, ${Down_{PHA}(\cdot )}$ and ${UP_{PHA}(\cdot )}$ stand for two linear layers, which down-projects the input to low rank, $r=16$, and up-projects it back to the rank of 768, respectively. Following the processing of ${F_{MSA}}$  through the MLP pathway, the output, ${F_{MLP}}$, serves as input to SHA and the detailed operation process can be performed as:
\begin{equation}\label{Eq10}
\begin{aligned}
F_{SHA} = UP_{SHA}(ReLU(Down_{SHA}(F_{MLP})))
\end{aligned}
\end{equation}
where ${F_{SHA}} \in \mathbb{R} {^{3N \times D}}$ represents the output of SHA, ${Down_{SHA}(\cdot )}$ and ${UP_{SHA}(\cdot )}$ are linear layers within SHA, akin to the other two in PHA. Finally, the multi-stage features from the three branches and the ${F_{MSA}}$  are fused to obtain the final feature $F$.
\begin{equation}\label{Eq11}
\begin{aligned}
F = F_{MSA}+F_{MLP}+F_{PHA}+F_{SHA}
\end{aligned}
\end{equation}

% =======
% FIG. 03
% =======

\subsection {Hyperspectral Enhancement of Input (HEI)}

To further exploit the spectral information in HS images, the HEI module is designed for early-stage processing. Different from HAS and HAM, which are utilized in feature extraction stages, HEI aims to integrate spectral information at the image level.

\begin{figure}
  \begin{center}
  \includegraphics[width=3in]{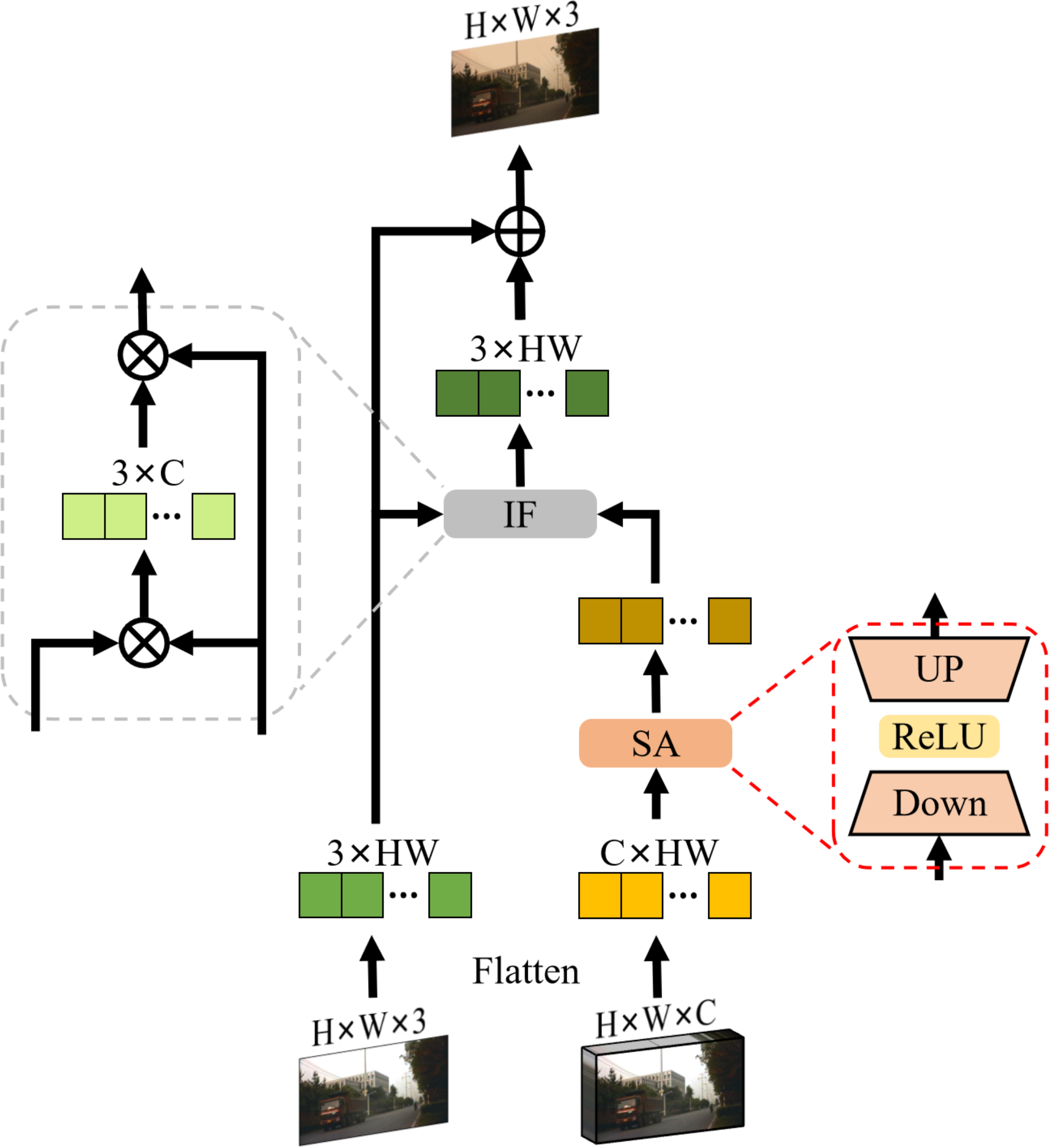}\\
 \caption{The structure of HEI. SA represents the spectral attention module. IF stands for the image-level fusion module.}\label{Fig4}
  \end{center}
\end{figure}

As the initial stage of the network, HEI takes the HS image and the corresponding false-color image from each of the three input pairs as input. For simplicity, the HEI is illustrated with the search pair, ${I_{S}^{HS}}$ and ${I_{S}^{FRGB}}$. The detailed architecture of HEI is presented in Fig.~\ref{Fig4}, comprising a Spectral Attention (SA) and an Image-level Fusion (IF) module.

Initially, HEI begins by transforming the two images ${I_{S}^{HS}}$ and ${I_{S}^{FRGB}}$ into a pair of flattened vectors, ${V_{HS}} \in \mathbb{R} {^{C \times HW}}$ and ${V_{FRGB}} \in \mathbb{R} {^{3 \times HW}}$.
\begin{equation}\label{Eq12}
\begin{aligned}
{V_{HS}} &= Flatten({I^{HS}}) \\
{V_{FRGB}} &= Flatten({I^{FRGB}})
\end{aligned}
\end{equation}
where $Flatten(\cdot )$ refers to the process of reshaping the input images into a one-dimensional sequence. In SA, the HS vector, ${V_{HS}}$, is processed through a set of low-rank learnable parameters, ${W_{down}} \in \mathbb{R} {^{3 \times C}}$ and ${W_{up}} \in \mathbb{R} {^{C \times 3}}$, with a ReLU activation interposed between them, resulting in an updated HS feature ${V_{SA}} \in \mathbb{R} {^{C \times HW}}$.The process in SA can be formulated as:
\begin{equation}\label{Eq13}
\begin{aligned}
V_{SA} = W_{up}(ReLU(W_{down}(V_{HS})))
\end{aligned}
\end{equation}
Subsequently, ${V_{SA}}$ is injected into the corresponding false-color vector ${V_{FRGB}}$, via the IF module through a parameter-free design, allowing information from the HS and false-color images to interact and inform each other.
\begin{equation}\label{Eq14}
\begin{aligned}
V_{E} = softmax({V_{FRGB}} \times {V_{SA}^{T}}) \times {V_{SA}}
\end{aligned}
\end{equation}
where $softmax(\cdot )$  represents the softmax activation function, and ${V_{E}}$ signifies the fused vector representation. To align with the input requirements of the feature extraction module, the enhanced vector ${V_{E}}$ is then mapped back to its corresponding image dimensions ${I^{E}} \in \mathbb{R} {^{H \times W \times 3}}$.

% =======
% FIG. 04
% =======

\subsection {Training and Inference}
\emph{Training.}
The HyA-T comprises two types of structures: those from the pretrained object tracking network and those designed for HS object tracking, i.e., HAS, HAM and HEI. The pretrained object tracking network is SeqTrack, which is a transformer-based tracking network and estimates the state of the target in a sequential manner \cite{43}. The parameters in the structures designed for HS object tracking are trained on HS datasets, while the parameters corresponding to SeqTrack are frozen. This method minimizes the number of parameters requiring adjustment, thereby expediting convergence. The loss function of the HyA-T utilizes the cross-entropy loss, and the details can be referred in \cite{43}.

\emph{Inference.}
The HyA-T captures the template from the initial frame in a sequence with the given information, and the search patches is obtained from subsequent images. The online template is initialized with the template and updated automatically when the confidence scores of the generated bounding boxes is higher than a certain threshold. The encoder in HyA-T jointly extracts features from the template, online template, and search patch, while the decoder exclusively utilizes the search patch features to generate the target's bounding box. For the object tracking task, the locations of the target in consecutive images are close to each other. Consequently, the Hanning window is applied on the score vector to penalize the location with large displacement.

% =======
% FIG. 04
% =======
% \begin{figure}
%   \begin{center}
%   \includegraphics[width=3in]{pdf-photo/HEI.pdf}\\
%  \caption{The structure of HEI. SA represents the spectral attention module. IF stands for the image-level fusion module.}\label{Fig4}
%   \end{center}
% \end{figure}

% === IV. Transistor Class-F inv Rectifier ========================================
% =================================================================================
\section{Experiments}

In this section, the implement details and datasets used for the experiments are introduced. Subsequently, the performance of HyA-T is evaluated in comparison with other SOTA trackers on HS object tracking datasets. Ablation studies are then conducted to assess the contributions of the proposed methods. Finally, the detailed analysis of attribute-based evaluations is provided, followed by the presentation of visualization results.

% =======
% Table. 01
% \usepackage[normalem]{ulem}
\begin{table*}[]
\centering
\setlength{\tabcolsep}{23.75pt} % 调节列间距
\begin{threeparttable} % 添加注释部分
\caption{Comparison with SOTA RGB models on HOTC dataset}
\label{table1} % 表格题目
\small
\renewcommand\arraystretch{1.2} % 控制行间距
\begin{tabular}{cccccc}
\toprule
                        &                          & \multicolumn{2}{c}{RGB}                                                      & \multicolumn{2}{c}{F-Color/HS}                                                 \\
\multirow{-2}{*}{Model} & \multirow{-2}{*}{Source} & AUC                                   & DP@20P                                & AUC                                   & DP@20P         \\ \midrule
SiamRPN \cite{29}                & CVPR 2018               & 0.592      & 0.902      & 0.486        & 0.757         \\
SiamPRN++ \cite{30}              & CVPR 2019               &{\color[HTML]{FF0000} \textbf{0.653}}      &{\color[HTML]{00B0F0} \textbf{0.912}}      & 0.591        & 0.847         \\
SiamDW \cite{61}                 & CVPR 2019               & 0.565      & 0.872      & 0.529        & 0.812         \\
SiamMask \cite{62}               & CVPR 2019               & 0.611      & 0.877      & 0.554        & 0.813         \\
PrDiMP \cite{63}                 & CVPR 2020               & 0.634      &{\color[HTML]{FF0000} \textbf{0.917}}      & 0.565        & 0.829         \\
SiamBAN \cite{64}                & CVPR 2020               & 0.610      & 0.853      & 0.587        &{\color[HTML]{00B0F0} \textbf{0.863}}         \\
SiamFC++ \cite{31}               & AAAI 2020               & 0.635      & 0.865      & 0.578        & 0.820         \\
SiamGAT \cite{65}                & CVPR 2021               &{\color[HTML]{00B0F0} \textbf{0.649}}      & 0.889      & 0.576        & 0.820         \\
STARK \cite{41}                  & ICCV 2021               & 0.637      & 0.900      & 0.579        & 0.814         \\
SiamCAR \cite{66}                & IJCV 2022               & 0.636      & 0.882      &{\color[HTML]{00B0F0} \textbf{0.586}}        & 0.846         \\
OSTrack \cite{42}                & ECCV 2022               & 0.621      & 0.897      & 0.558        & 0.818         \\
SeqTrack \cite{43}               & CVPR 2023               & 0.612      & 0.890      & 0.583        & 0.860         \\
SMAT \cite{67}                   & WACV 2024               & 0.637      & 0.894      & 0.581        & 0.831         \\
HyA-T                  & Ours                    & -          & -          & {\color[HTML]{FF0000} \textbf{0.705}}        & {\color[HTML]{FF0000} \textbf{0.958}}        \\ \bottomrule
\end{tabular}
\begin{tablenotes} % 添加注释部分
\item The red and blue are utilized to mark the top two values. % 注释内容
\end{tablenotes} % 注释结束
\end{threeparttable} % threeparttable 结束
\end{table*}

\subsection {Experimental Setup}

\subsubsection{\textbf{Implementation Details}} A server equipping with one Intel Xeon Silver 4210R CPU and two NVIDIA RTX 3090 GPUs is utilized in the experiments. HyA-T is implemented in Python using the PyTorch framework. The autoregressive decoder is adapted from SeqTrack \cite{43}, while the feature extraction network is based on a modified version of the ViT \cite{23}. During training, the parameters of both the autoregressive decoder and the ViT-based feature extraction network are frozen. In HyA-T, the HEI and HAS components are trained with a learning rate of 1e-4, while the HAM is trained with a learning rate of 1e-5. A weight decay of 1e-4 is applied, and the AdamW optimizer \cite{60} is used with a batch size of 24. The total training epoch is set to 30 and the learning rate reduced by a factor of 10 after 24 epochs. The input images, i.e., the template, online template, and search images, are all set to 256×256 pixels. Additionally, the reduced channel size in the adapter is set to 16.

\subsubsection{\textbf{Dataset}} Four datasets are applied in the experiments, and the details of these datasets are presented as follows.

\emph{HOTC Dataset:}
HOTC \cite{1} is the first public HS tracking dataset in the Hyperspectral Object Tracking Challenge 2022. The dataset comprises three type of videos, i.e., HS videos, false-color videos and RGB videos. Each frame in the HS videos consists of 16 spectral bands. The false-color videos are generated by converting the HS videos with the CIE \cite{1} method, while the RGB videos are captured from a similar viewpoint as the HS videos. The dataset contains 40 videos for training and 35 videos for testing. Each videos in the HOTC dataset is annotated with 11 attributes to represent different challenging scenarios: Out-of-View (OV), Low Resolution (LR), Background Clutter (BC), Motion Blur (MB), Fast Motion (FM), In-Plane Rotation (IPR), Out-of-Plane Rotation (OPR), Scale Variation (SV), Illumination Variation (IV), Occlusion (OCC) and Deformation (DEF).

% =======
% Table. 04
% \usepackage{multirow}
% \begin{table*}[]
% \centering
% \setlength{\tabcolsep}{31pt} % 调节列间距
% \renewcommand\arraystretch{1.2} % 控制行间距
% \begin{threeparttable} % 添加注释部分
% \caption{Comparison of AUC and DP@20P with different configurations}
% \label{table4} % 表格题目
% \small
% \begin{tabular}{llllll}
% \toprule
%                         & AUC        & $\Delta$AUC & DP@20P   & $\Delta$DP@20P \\ \midrule
% Baseline                & 0.365      & -           & 0.454    & -              \\
% +HEI                    & 0.385      & 0.020       & 0.492    & 0.038          \\
% +HEI+HAS                & 0.489      & 0.124       & 0.652    & 0.198          \\
% +HEI+HAS+HAM            & {\color[HTML]{FF0000} \textbf{0.546}}      & 0.181       & {\color[HTML]{FF0000} \textbf{0.689}}    & 0.235          \\ \bottomrule
% \end{tabular}
% \begin{tablenotes} % 添加注释部分
% \item The red is utilized to mark the top value. % 注释内容
% \end{tablenotes} % 注释结束
% \end{threeparttable} % threeparttable 结束
% \end{table*}

\emph{HOTC2024 Dataset:}
The HOTC2024-VIS \cite{24}, HOTC2024-NIR \cite{24}, HOTC2024-RedNIR \cite{24} are introduced in the Hyperspectral Object Tracking Challenge 2024. The data in HOTC2024-VIS, HOTC2024-NIR, HOTC2024-RedNIR comprise 16, 25, and 15 spectral bands, respectively. Similar to the HOTC dataset, the HOTC2024 datasets include the false-color videos corresponding to the HS videos which are generated by the CIE. The HOTC2024-VIS provides 111 videos for training and 67 videos for testing, the HOTC2024-NIR provides 70 videos for training and 30 videos for testing, and the HOTC2024-RedNIR provides 36 videos for training and 20 videos for testing.

\subsection {Comparison on HOTC Dataset}
To demonstrate the effectiveness of the proposed tracker, extensive experiments are conducted on the HOTC dataset \cite{1}. The comparison results between HyA-T and SOTA RGB trackers are presented in Table \ref{table1}, while the comparison results between HyA-T and SOTA HS trackers are shown in Table \ref{table2}. A detailed analysis is provided as follows.

% =======
% Table. 02
% =======
\begin{table*}[]
\centering
\setlength{\tabcolsep}{6.5pt} % 调节列间距
\renewcommand\arraystretch{1.5} % 控制行间距
\begin{threeparttable}
\caption{comparison with SOTA HS models on HOTC dataset}
\label{table2}
\small
\begin{tabular}{cccccccc}
\hline
Model  & MHT \cite{1}     & MFI-HVT \cite{68} & BAE-Net \cite{69} & SiamHT \cite{12} & SSDT-Net \cite{50} & SiamBAG \cite{17} & TBR-Net \cite{22} \\ \hline
AUC    & 0.586   & 0.601   & 0.606   & 0.621  & 0.639    & 0.641   & 0.660   \\
DP@20P & 0.882   & 0.893   & 0.878   & 0.878  & 0.916    & 0.904   & 0.920   \\ \hline
Model  & PHTrack \cite{21} & SEE-Net \cite{14} & SPIRIT \cite{15}  & SENSE \cite{70}  & MMF-Net \cite{6}  & HyA-T   &         \\ \hline
AUC    & 0.660   & 0.666   & 0.679   & 0.690  &{\color[HTML]{00B0F0} \textbf{0.691}}    &{\color[HTML]{FF0000} \textbf{0.705}}   &         \\
DP@20P & 0.919   & 0.932   & 0.925   &{\color[HTML]{00B0F0} \textbf{0.952}}  & 0.932    &{\color[HTML]{FF0000} \textbf{0.958}}   &         \\ \hline
\end{tabular}
\begin{tablenotes} % 添加注释部分
\item The red and blue are utilized to mark the top two values. % 注释内容
\end{tablenotes} % 注释结束
\end{threeparttable}
\end{table*}

% =======
% Table. 03
% \usepackage{multirow}
\begin{table*}[]
\centering
\setlength{\tabcolsep}{13.25pt} % 调节列间距
\renewcommand\arraystretch{1.2} % 控制行间距
\begin{threeparttable} % 添加注释部分
\caption{comparison with SOTA models on HOTC2024-VIS, HOTC2024-NIR, and HOTC2024-RedNIR datasets}
\label{table3} % 表格题目
\small
\begin{tabular}{cccccccc}
\toprule
                        &                          & \multicolumn{2}{c}{HOTC2024-VIS}                                                      & \multicolumn{2}{c}{HOTC2024-NIR}                                                 & \multicolumn{2}{c}{HOTC2024-RedNIR}                                                 \\
\multirow{-2}{*}{Model} & \multirow{-2}{*}{Source} & AUC                                   & DP@20P                                & AUC                                   & DP@20P         & AUC         & DP@20P        \\ \midrule
SEE-Net \cite{14}                & TIP 2023                & 0.396      & 0.560      & 0.509      & 0.769      & 0.383       & 0.521        \\
Trans-DAT \cite{71}               & TCSVT 2024              & 0.397      & 0.524      & 0.587      & 0.753      &{\color[HTML]{00B0F0} \textbf{0.423}}       &{\color[HTML]{00B0F0} \textbf{0.547}}        \\
SPIRIT \cite{15}                 & TGRS 2024               & 0.319      & 0.409      & 0.655      & 0.823      & 0.377       & 0.516        \\
MMF-Net \cite{6}                & TGRS 2024               &{\color[HTML]{00B0F0} \textbf{0.482}}      &{\color[HTML]{00B0F0} \textbf{0.645}}      &{\color[HTML]{00B0F0} \textbf{0.701}}      &{\color[HTML]{00B0F0} \textbf{0.875}}      & 0.388       & 0.521        \\
PHTrack \cite{21}                & TGRS 2024               & 0.303      & 0.410      & 0.526      & 0.731      & 0.262       & 0.364        \\
SENSE \cite{70}                  & Inform Fusion 2024      & 0.298      & 0.398      & 0.561      & 0.765      & 0.364       & 0.465        \\
HyA-T                  & Ours                    & {\color[HTML]{FF0000} \textbf{0.554}}      & {\color[HTML]{FF0000} \textbf{0.712}}      & {\color[HTML]{FF0000} \textbf{0.776}}      & {\color[HTML]{FF0000} \textbf{0.935}}      & {\color[HTML]{FF0000} \textbf{0.546}}       & {\color[HTML]{FF0000} \textbf{0.689}}        \\ \bottomrule
\end{tabular}
\begin{tablenotes} % 添加注释部分
\item The red and blue are utilized to mark the top two values. % 注释内容
\end{tablenotes} % 注释结束
\end{threeparttable} % threeparttable 结束
\end{table*}

\subsubsection{\textbf{Comparison with RGB Trackers}}
Table \ref{table1} presents the experimental results of HyA-T and other 13 RGB trackers, i.e., SiamRPN \cite{29}, SiamRPN++ \cite{30}, SiamDW \cite{61}, SiamMask \cite{62}, PrDiMP \cite{63}, SiamBAN \cite{64}, SiamFC++ \cite{31}, SiamGAT \cite{65}, STARK \cite{41}, SiamCAR \cite{66}, OSTrack \cite{42}, SeqTrack \cite{43} and SMAT \cite{67}. The performance of HyA-T is evaluated on HS videos, and the SOTA RGB trackers are evaluated on both RGB and false-color videos.

From the results in Table \ref{table1}, it is evident that RGB trackers perform significantly better on RGB videos than on false-color videos, highlighting their reduced effectiveness in handling the spectral differences of false-color representations. This differences emphasize the importance of designing trackers specifically for tracking the target HS videos. As shown in Table \ref{table1}, HyA-T achieves the best performance with an AUC score of 0.705 and a DP@20P score of 0.958. Compared with the second best methods, i.e., SiamRPN++ and PrDiMP, HyA-T achieves significantly enhancements of 5.2$\%$ on AUC and 4.1$\%$ on DP@20P scores. For the experiments conducted on false-color videos, HyA-T outperforms the SiamCAR by 11.9$\%$ on AUC, and SiamBAN by 9.5$\%$ on DP@20P.

\subsubsection{\textbf{Comparison with Hyperspectral Trackers}}
Table \ref{table2} presents the AUC and DP@20P scores for HyA-T and 12 SOTA HS trackers, i.e., MHT \cite{1}, MFI-HVT \cite{68}, BAE-Net \cite{69}, SiamHT \cite{12}, SSDT-Net \cite{50}, SiamBAG \cite{17}, TBR-Net \cite{22}, PHTrack \cite{21}, SEE-Net \cite{14}, SPIRIT \cite{15}, SENSE \cite{70}, and MMF-Net \cite{6}. Among the compared SOTA HS trackers, MHT and MFI-HVT, which rely on handcrafted features, exhibit weaker tracking performance. In contrast, the other trackers using deep features result in better performance. SiamHT and SSDT-Net, which convert the HS image into single three-channel image, shows worse performance due to the loss of spectral information. BAE-Net, SiamBAG, TBR-Net, PHTrack, SEE-Net, SPIRIT, and SENSE preprocess the input HS image by regrouping it into multiple three-channel images, and track the target based on these images. The limited performances of these trackers attribute to the loss of interactive information between the spectral bands. SENSE achieves a DP@20P score of 0.952, ranking second in Table \ref{table2}. But HyA-T outperforms SENSE 1.5$\%$ on AUC and 0.6$\%$ on DP@20P. MMF-Net employs an unmixing method to generate multiple data sources, incorporating different information to achieve accurate tracking, and ranks second in AUC. However, HyA-T still outperforms MMF-Net by 1.4$\%$ on AUC and 2.6$\%$ on DP@20P. As shown in Table \ref{table2}, HyA-T achieves superior performance in both AUC and DP@20P scores. The impressive tracking performance can be attributed to the proposed modules, which enhance the tracker’s ability to process HS information.

\subsection {Comparison on HOTC2024 Dataset}

To further confirm the superiority of HyA-T, comprehensive experiments are conducted on the HOTC2024 dataset \cite{24}. Table \ref{table3} presents the tracking performance of HyA-T and other seven SOTA HS trackers, i.e., SEE-Net \cite{14}, Trans-DAT \cite{71}, SPIRIT \cite{15}, MMF-Net \cite{6}, PHTrack \cite{21}, SENSE \cite{70}, and SSTtrack \cite{47}.

% =======
% Table. 05
% \usepackage{multirow}
% \begin{table*}[]
% \centering
% \setlength{\tabcolsep}{30pt} % 调节列间距
% \renewcommand\arraystretch{1.2} % 控制行间距
% \begin{threeparttable} % 添加注释部分
% \caption{ablation study of HEI on HOTC2024-RedNIR dataset}
% \label{table5} % 表格题目
% \small
% \begin{tabular}{cccccc}
% \toprule
%                         & AUC        & $\Delta$AUC & DP@20P   & $\Delta$DP@20P \\ \midrule
% Baseline                & 0.365      & -           & 0.454    & -              \\
% Concat\&Downsample      & 0.266      & -0.099      & 0.357    & -0.097         \\
% Downsample\&Add         & 0.356      & -0.009      & 0.454    & 0              \\
% SA\&Downsample\&Add     & 0.382      & 0.017       & 0.473    & 0.019          \\
% HEI                     & {\color[HTML]{FF0000} \textbf{0.385}}      & 0.020       & {\color[HTML]{FF0000} \textbf{0.492}}    & 0.038          \\ \bottomrule
% \end{tabular}
% \begin{tablenotes} % 添加注释部分
% \item The red is utilized to mark the top value. % 注释内容
% \end{tablenotes} % 注释结束
% \end{threeparttable} % threeparttable 结束
% \end{table*}

% =======
% Table. 04
% \usepackage{multirow}
\begin{table}[htbp]
\centering
\setlength{\tabcolsep}{5.75pt} % 调节列间距
\renewcommand\arraystretch{1.2} % 控制行间距
\begin{threeparttable} % 添加注释部分
\caption{ablation study of HyA-T on HOTC2024-RedNIR dataset}
\label{table4} % 表格题目
\small
\begin{tabular}{lccccc}
\toprule
                        & AUC        & $\Delta$AUC & DP@20P   & $\Delta$DP@20P \\ \midrule
Baseline                & 0.365      & -           & 0.454    & -              \\
+HEI                    & 0.385      & 0.020       & 0.492    & 0.038          \\
+HEI+HAS                & 0.489      & 0.124       & 0.652    & 0.198          \\
+HEI+HAS+HAM            & {\color[HTML]{FF0000} \textbf{0.546}}      & 0.181       & {\color[HTML]{FF0000} \textbf{0.689}}    & 0.235          \\ \bottomrule
\end{tabular}
\begin{tablenotes} % 添加注释部分
\item The red is utilized to mark the top value. % 注释内容
\end{tablenotes} % 注释结束
\end{threeparttable} % threeparttable 结束
\end{table}

% =======
% Table. 05
% \usepackage{multirow}
\begin{table}[htbp]
\centering
\setlength{\tabcolsep}{4pt} % 调节列间距
\renewcommand\arraystretch{1.2} % 控制行间距
\begin{threeparttable} % 添加注释部分
\caption{ablation study of HEI on HOTC2024-RedNIR dataset}
\label{table5} % 表格题目
\small
\begin{tabular}{cccccc}
\toprule
                        & AUC        & $\Delta$AUC & DP@20P   & $\Delta$DP@20P \\ \midrule
Baseline                & 0.365      & -           & 0.454    & -              \\
Concat\&Downsample      & 0.266      & -0.099      & 0.357    & -0.097         \\
Downsample\&Add         & 0.356      & -0.009      & 0.454    & 0              \\
SA\&Downsample\&Add     & 0.382      & 0.017       & 0.473    & 0.019          \\
HEI                     & {\color[HTML]{FF0000} \textbf{0.385}}      & 0.020       & {\color[HTML]{FF0000} \textbf{0.492}}    & 0.038          \\ \bottomrule
\end{tabular}
\begin{tablenotes} % 添加注释部分
\item The red is utilized to mark the top value. % 注释内容
\end{tablenotes} % 注释结束
\end{threeparttable} % threeparttable 结束
\end{table}

\begin{figure*}[!t]
\centering
\subfloat{
    \label{Fig5_1}\includegraphics[width=3in]{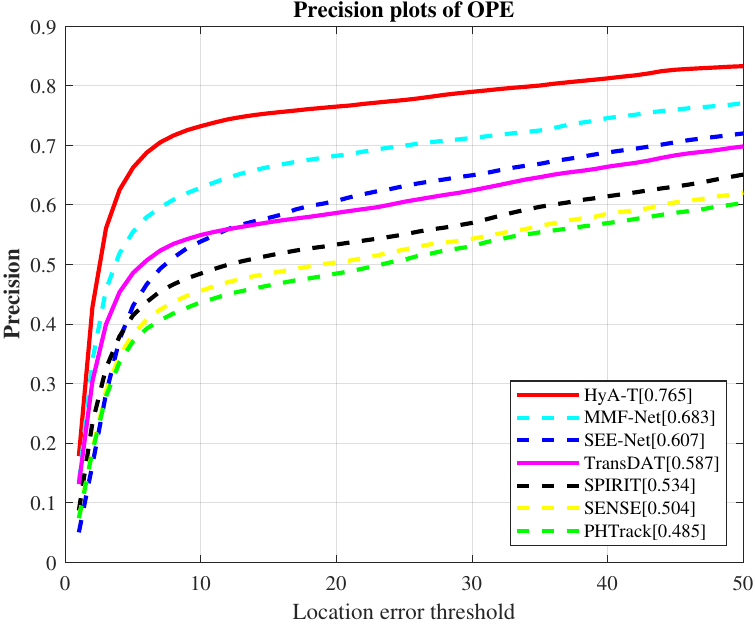}
}
\subfloat{
    \label{Fig5_2}\includegraphics[width=3in]{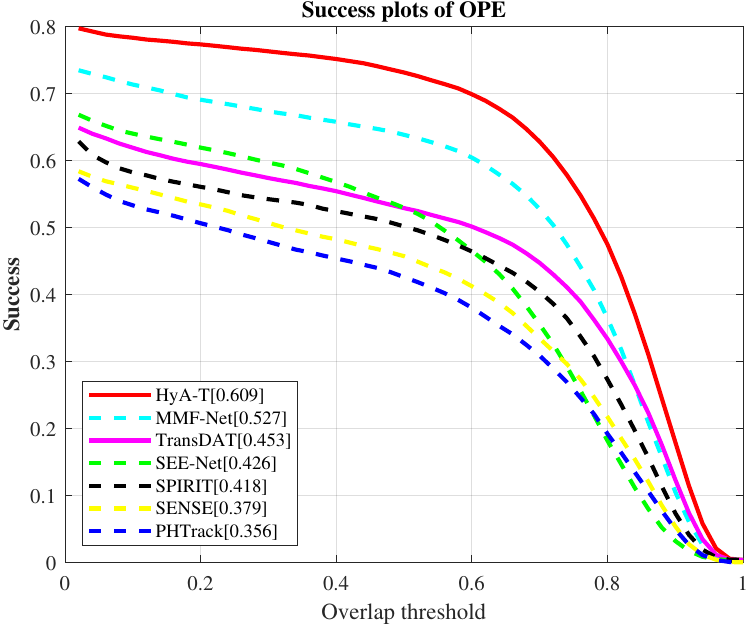}
}
\caption{Comparisons of HyA-T and other SOTA HS trackers on the HOTC2024 dataset.}
\label{Fig5}
\end{figure*}

\subsubsection{\textbf{HOTC2024-VIS}}

HOTC2024-VIS consists of 67 test videos, and each video contains HS images with 16 spectral bands. As shown in Table \ref{table3}, HyA-T achieves the best performance, 0.554 on AUC and 0.712 on DP@20P. Compared to the second tracker, MMF-Net, HyA-T shows a significant improvement, with a gain of 7.2$\%$ on AUC and 6.7$\%$ on DP@20P, respectively.

\subsubsection{\textbf{HOTC2024-NIR}}

HOTC2024-NIR includes 30 test videos for NIR data, which contains 25 spectral bands. As shown in Table \ref{table3}, HyA-T ranks first on AUC score with values of 0.776, and on DP@20P score with values of 0.935. Compared to the second tracker, MMF-Net, the results clearly indicate that the proposed modules make HyA-T achieve outstanding performance, surpassing MMF-Net by 7.5$\%$ on AUC and 6.0$\%$ on DP@20P, respectively.

\subsubsection{\textbf{HOTC2024-RedNIR}}

HOTC2024-RedNIR has 15 spectral bands in HS images, providing 20 videos for testing. As shown in Table \ref{table3}, HyA-T achieves the best performance with AUC and DP@20P scores of 0.546 and 0.689, respectively. HyA-T significantly outperforms Trans-DAT by 12.3$\%$ on AUC and 14.2$\%$ on DP@20P, respectively.

\subsubsection{\textbf{Overall}}

Fig.~\ref{Fig5} illustrates the overall tracking performance on the HOTC2024 dataset, showing the success and precision plots for the trackers in Table \ref{table3}. HyA-T achieves superior overall performance with an AUC score of 0.609 and a DP@20P score of 0.765. It demonstrates a significant advantage over other trackers, especially MMF-Net, with improvements of 8.2$\%$ on AUC and DP@20P scores, respectively.

The analysis above clearly demonstrates the superior performance of HyA-T and the effectiveness of the proposed modules.

\subsection {Ablation Study}

\subsubsection{\textbf{Effectiveness of individual proposed method}}

In this section, the contribution of each proposed method to the tracking performance is evaluated on the HOTC2024-RedNIR \cite{24}. The results are presented in Table \ref{table4}. The Baseline is SeqTrack, the foundational network for HyA-T, which is fine-tuned on the HOTC2024-RedNIR dataset. As shown in the comparison between the first and second lines, incorporating the HEI improves the performance by 2.0$\%$ on AUC and 3.8$\%$ on DP@20P, which verifies that augments the spectral information into the network inputs is effective. Additionally, the HAS, which adapts the tracking network to HS data, provides a substantial performance boost, improving AUC by 10.4$\%$ and DP@20P by 16.0$\%$. The modification of the MLP in the encoder layers, termed HAM, further enhances performance of the tracker by 5.7$\%$ on AUC and 3.7$\%$ on DP@20P. In general, the proposed methods adapt the base tracking network to the HS object tracking task with the spectral information, and improve the performance noticeably, i.e., 17.2$\%$ on AUC and 23.5$\%$ on DP@20P.

% =======
% Table. 06
% \usepackage{multirow}
\begin{table*}[]
\centering
\setlength{\tabcolsep}{19.75pt} % 调节列间距
\renewcommand\arraystretch{1.2} % 控制行间距
\begin{threeparttable} % 添加注释部分
\caption{ablation study of HAS on HOTC2024-RedNIR dataset}
\label{table6} % 表格题目
\small
\begin{tabular}{cccccccc}
\toprule
Index & Query & Key & Value & AUC   & $\Delta$AUC & DP@20P & $\Delta$DP@20P \\ \midrule
1     & -     & -   & -     & 0.385 & -           & 0.492  & -              \\
2     & \checkmark     & -   & -     & 0.449 & 0.064       & 0.568  & 0.076          \\
3     & -     & \checkmark   & -     & 0.487 & 0.102       & 0.623  & 0.131          \\
4     & -     & -   & \checkmark     & 0.477 & 0.092       & 0.625  & 0.133          \\
5     & \checkmark     & \checkmark   & -     & 0.456 & 0.071       & 0.574  & 0.082          \\
6     & \checkmark     & -   & \checkmark     & {\color[HTML]{FF0000} \textbf{0.489}} & 0.104       & {\color[HTML]{FF0000} \textbf{0.652}}  & 0.160          \\
7     & -     & \checkmark   & \checkmark     & 0.479 & 0.094       & 0.603  & 0.111          \\
8     & \checkmark     & \checkmark   & \checkmark     & 0.471 & 0.086       & 0.605  & 0.113          \\ \bottomrule
\end{tabular}
\begin{tablenotes} % 添加注释部分
\item The red is utilized to mark the top value. % 注释内容
\end{tablenotes} % 注释结束
\end{threeparttable} % threeparttable 结束
\end{table*}

% =======
% Table. 07
% \usepackage{multirow}
\begin{table}[htbp]
\centering
\setlength{\tabcolsep}{2pt} % 调节列间距
\renewcommand\arraystretch{1.2} % 控制行间距
\begin{threeparttable} % 添加注释部分
\caption{ablation study of Linear Projection on HOTC2024-RedNIR dataset}
\label{table7} % 表格题目
\small
\begin{tabular}{cccccc}
\toprule
                          & AUC        & $\Delta$AUC & DP@20P   & $\Delta$DP@20P \\ \midrule
Single False-Color Image  & 0.462      & -           & 0.598    & -              \\
Multiple False-Color Images & 0.469     & 0.007       & 0.601    & 0.003          \\
Original Hyperspectral Image & {\color[HTML]{FF0000} \textbf{0.489}}    & 0.027       & {\color[HTML]{FF0000} \textbf{0.652}}    & 0.054          \\ \bottomrule
\end{tabular}
\begin{tablenotes} % 添加注释部分
\item The red is utilized to mark the top value. % 注释内容
\end{tablenotes} % 注释结束
\end{threeparttable} % threeparttable 结束
\end{table}

\subsubsection{\textbf{Design of the HEI}}

In this section, the different methods for augmenting spectral information into the inputs of the tracking network are evaluated. Table \ref{table5} presents the results. The Baseline is the same method from Table \ref{table4}. The Concat$\&$Downsample concatenates the HS image with its corresponding false-color image along the channel dimension, followed by dimensionality reduction into a three-channel image with a convolution layer. The Downsample$\&$Add reduces the HS image into the three-channel image with a convolution layer and then adds it to the false-color image via element-wise summation. The SA$\&$Downsample$\&$Add is similar with the Downsample$\&$Add, which incorporates the SA module to update the HS image before applying the same dimensionality reduction and summation process.

As shown in Table \ref{table5}, the Concat$\&$Downsample performs the worst due to significant spectral information loss during dimensionality reduction. The Downsample$\&$Add yields similar results to the Baseline. The SA$\&$Downsample$\&$Add utilizes the SA to enhance the spectral information in the HS image and results in a slight improvement. However, the dimensionality reduction of the HS image and the simple element-wise summation distort the information in the input image. In contrast, the HEI augments the spectral information in the original HS image into the inputs, and gains improvement of 2.0$\%$ on AUC and 3.8$\%$ on DP@20P.

% \begin{figure}[!t]
% \centering
% \subfloat{
%     \label{Fig5_1}\includegraphics[width=1.7in]{pdf-photo/precision.pdf}
% }
% \subfloat{
%     \label{Fig5_2}\includegraphics[width=1.7in]{pdf-photo/AUC.pdf}
% }
% \caption{Comparisons of HyA-T and other SOTA HS trackers on the HOTC2024 dataset.}
% \label{Fig5}
% % \vspace{-10pt}
% \end{figure}

\subsubsection{\textbf{Analysis on HAS}}

Based on the tracker equipping the HEI, the HAS is applied on different set of queries, keys and values in the self-attention learning within the encoder layer. The results are reported in Table \ref{table6}. When the HAS is applied to the queries, keys, and values individually, improvements in tracker performance are observed, as shown in the second, third, and fourth lines of Table \ref{table6}. Furthermore, applying the HAS to various combinations of queries, keys, and values (lines five to eight) also leads to performance enhancements. Among these combinations, the method in the sixth line achieves the best performance across all approaches in Table \ref{table6}. When HAS is applied only to the keys (third line), the AUC score is comparable to that of the method in the sixth line. However, the DP@20P score is slightly lower than the best-performing combination, which applies HAS to both the queries and the values.

Moreover, to evaluate the generation of adapter in the HAS with different strategies for processing HS images, a experiment is conducted and the results are reported in Table \ref{table7}. The HAS extracts the spectral information directly from the original HS images, and the other existing HS object tracking methods extracts the spectral information from single or multiple false-color images. Hence, the HAS learning the spectral information from single false-color image, multiple false-color images and the original HS images are evaluated in this experiment. Specifically, the method in the first line divides the HS image into several false-color images, which are then fused into a single false-color image through element-wise addition. The method in the second line also divided the HS image into several false-color images, and learns the spectral information from these false-color images. The method in the third line extracts the spectral information directly from the original HS image. As shown in Table \ref{table7}, the method in the second line outperforms the method in the first line since more spectral information is reserved. However, it performs worse than the method in the third line due to the loss of the interactive information. Consequently, the HAS learns the spectral information directly from the original HS images.

% % =======
% % Table. 07
% % =======
% % Please add the following required packages to your document preamble:
% % \usepackage{multirow}
% % \usepackage[table,xcdraw]{xcolor}
% % Beamer presentation requires \usepackage{colortbl} instead of \usepackage[table,xcdraw]{xcolor}
% \begin{table}[htbp]
% \centering
% \setlength{\tabcolsep}{2pt} % 调节列间距
% \renewcommand\arraystretch{1.2} % 控制行间距
% \begin{threeparttable} % 添加注释部分
% \caption{Ablation study results}
% \label{table7} % 表格题目
% \small
% \begin{tabular}{cccccc}
% \toprule
% Method                       & AUC   & $\Delta$AUC  & DP@20P & $\Delta$DP@20P \\ \midrule
% Single False-Color Image     & 0.462 & -            & 0.598  & -              \\
% Multiple False-Color Images  & 0.469 & 0.007        & 0.601  & 0.003          \\
% Original Hyperspectral Image & 0.489 & 0.027        & 0.652  & 0.054          \\ \bottomrule
% \end{tabular}
% \begin{tablenotes} % 添加注释部分
% \item The table shows the performance of different methods. % 注释内容
% \end{tablenotes} % 注释结束
% \end{threeparttable} % threeparttable 结束
% \end{table}

\subsubsection{\textbf{Design of the HAM}}

In this part, the evaluation of different implementations of the HAM is conducted, with results presented in Table \ref{table8}. Seq. and Par. represent the sequential and parallel establishment of the HAM, respectively. Both the sequential and parallel implementations enhance the performance by adapting the MLP to the HS object tracking task. Furthermore, the simultaneous use of both sequential and parallel implementations of the HAM results in an additional performance improvement.

% % =======
% % Table. 08
% % =======
\begin{table*}[]
\centering
\setlength{\tabcolsep}{30pt} % 调节列间距
\renewcommand\arraystretch{1.2} % 控制行间距
\begin{threeparttable} % 添加注释部分
\caption{ablation study of HAM on HOTC2024-RedNIR dataset}
\label{table8} % 表格题目
\small
\begin{tabular}{cccccc}
\toprule
Seq. & Par. & AUC   & $\Delta$AUC  & DP@20P & $\Delta$DP@20P \\ \midrule
-    & -    & 0.489 & -            & 0.652  & -              \\
\checkmark    & -    & 0.517 & 0.028        & 0.666  & 0.014          \\
-    & \checkmark    & 0.518 & 0.029        & 0.671  & 0.019          \\
\checkmark    & \checkmark    & {\color[HTML]{FF0000} \textbf{0.546}} & 0.057 & {\color[HTML]{FF0000} \textbf{0.689}} & 0.037 \\ \bottomrule
\end{tabular}
\begin{tablenotes} % 添加注释部分
\item The red is utilized to mark the top value. % 注释内容
\end{tablenotes} % 注释结束
\end{threeparttable} % threeparttable 结束
\end{table*}

% % =======
% % Table. 09
% % =======
\begin{table*}[]
\centering
\setlength{\tabcolsep}{4.25pt} % 调节列间距
\renewcommand\arraystretch{1.2} % 控制行间距
\begin{threeparttable} % 添加注释部分
\caption{attributes-based comparison with SOTA HS trackers on HOTC dataset}
\label{table9} % 表格题目
\small
\begin{tabular}{ccccccccccc}
\toprule
Attributes & MHT  & BAE-Net & SSDT-Net & SiamBAG & PHTrack & SEE-Net & SPIRIT & SENSE & MMF-Net & HyA-T \\ \midrule
Background Clutter (BC)         & 0.606 & 0.651   & 0.663    & 0.648   & 0.678   & 0.705   & 0.715  &{\color[HTML]{FF0000} \textbf{0.721}} & 0.708   &{\color[HTML]{00B0F0} \textbf{0.716}} \\
Deformation (DEF)        & 0.688 & 0.709   & 0.685    & 0.691   & 0.705   & 0.742   & 0.756  & 0.722 &{\color[HTML]{00B0F0} \textbf{0.761}}   &{\color[HTML]{FF0000} \textbf{0.763}} \\
Fast Motion (FM)         & 0.542 & 0.608   & 0.603    & 0.615   & 0.735   & 0.711   & 0.715  &{\color[HTML]{00B0F0} \textbf{0.745}} & 0.704   &{\color[HTML]{FF0000} \textbf{0.746}} \\
In-Plane Rotation (IPR)        & 0.658 & 0.654   & 0.666    & 0.693   & 0.710   & 0.725   & 0.752  & 0.743 &{\color[HTML]{00B0F0} \textbf{0.766}}   &{\color[HTML]{FF0000} \textbf{0.768}} \\
Illumination Variation (IV)         & 0.491 & 0.507   & 0.543    & 0.529   & 0.556   & 0.541   & 0.534  &{\color[HTML]{00B0F0} \textbf{0.592}} & 0.572   &{\color[HTML]{FF0000} \textbf{0.624}} \\
Low Resolution (LR)         & 0.462 & 0.572   & 0.521    & 0.567   & 0.614   &{\color[HTML]{FF0000} \textbf{0.672}}   & 0.647  & 0.620 &{\color[HTML]{00B0F0} \textbf{0.665}}   & 0.628 \\
Motion Blur (MB)         & 0.616 & 0.629   & 0.579    & 0.680   & 0.733   & 0.700   & 0.726  & 0.732 &{\color[HTML]{FF0000} \textbf{0.773}}   &{\color[HTML]{00B0F0} \textbf{0.745}} \\
Occlusion (OCC)        & 0.561 & 0.521   & 0.607    & 0.592   & 0.591   & 0.605   & 0.624  & 0.623 &{\color[HTML]{00B0F0} \textbf{0.631}}   &{\color[HTML]{FF0000} \textbf{0.663}} \\
Out-of-Plane Rotation (OPR)        & 0.635 & 0.659   & 0.695    & 0.674   & 0.696   & 0.714   & 0.745  & 0.729 &{\color[HTML]{00B0F0} \textbf{0.755}}   &{\color[HTML]{FF0000} \textbf{0.770}} \\
Out-of-View (OV)         & 0.423 & 0.598   &{\color[HTML]{00B0F0} \textbf{0.732}}    & 0.554   & 0.665   & 0.664   & 0.698  & 0.677 & 0.671   &{\color[HTML]{FF0000} \textbf{0.768}} \\
Scale Variation (SV)         & 0.574 & 0.600   & 0.639    & 0.619   & 0.644   & 0.633   & 0.651  & 0.666 &{\color[HTML]{00B0F0} \textbf{0.673}}   &{\color[HTML]{FF0000} \textbf{0.683}} \\ \bottomrule
\end{tabular}
\begin{tablenotes} % 添加注释部分
\item The red and blue are utilized to mark the top two values. % 注释内容
\end{tablenotes} % 注释结束
\end{threeparttable} % threeparttable 结束
\end{table*}

% =======
% FIG. 06
% =======
\begin{figure*}
  \begin{center}
  \includegraphics[width=7.25in]{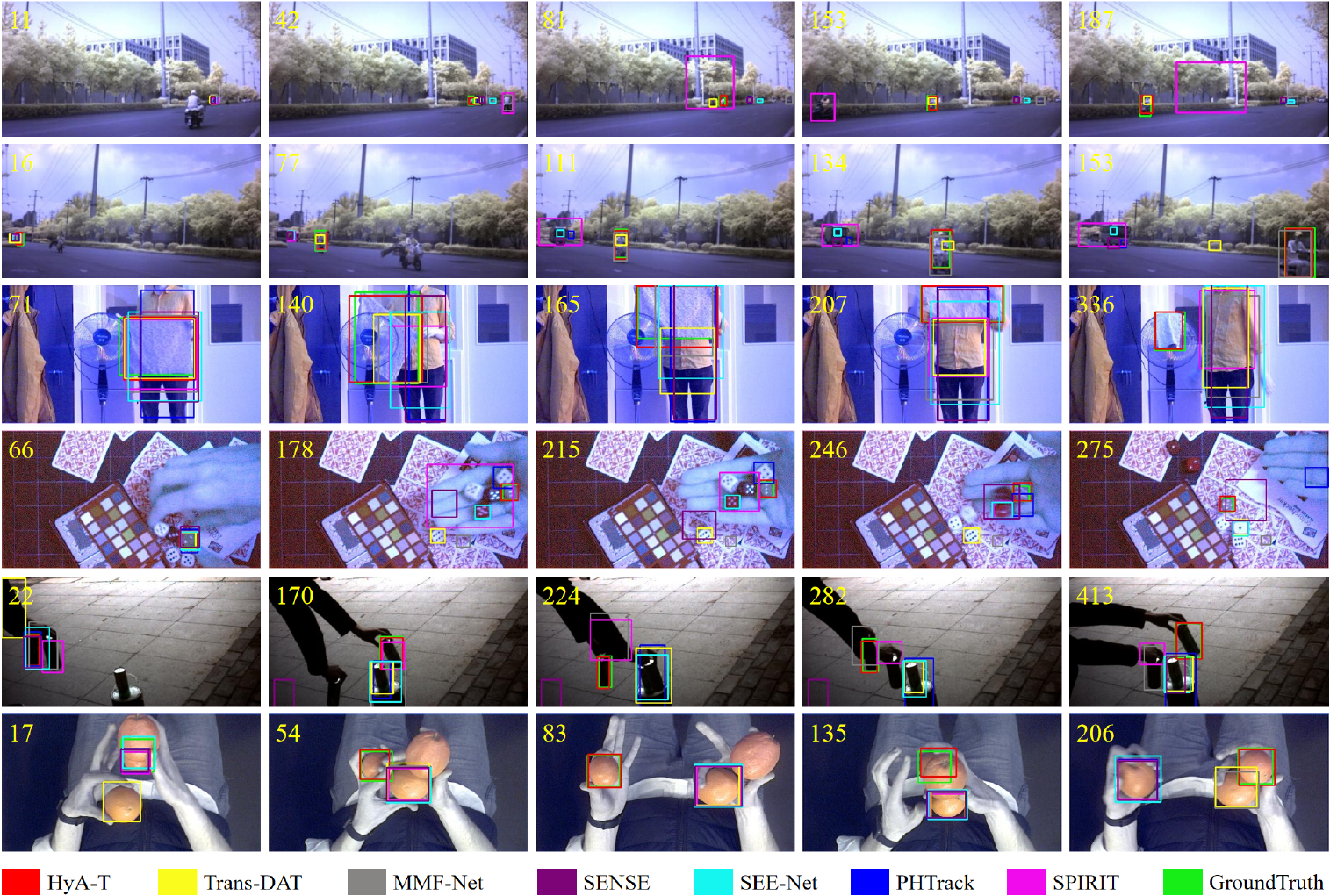}\\
 \caption{Visualization comparisons of HyA-T and other SOTA HS trackers on the HOTC2024 dataset.}\label{Fig6}
  \end{center}
\end{figure*}

\subsection {Attributes-based Evaluation}

To comprehensively assess the tracking ability of HyA-T, extensive experiments are conducted across 11 attributes on the HOTC dataset. Table \ref{table9} presents the attribute-based comparison results in terms of AUC scores between HyA-T and nine other HS trackers: MHT \cite{1}, BAE-Net \cite{69}, SSDT-Net \cite{50}, SiamBAG \cite{17}, PHTrack \cite{21}, SEE-Net \cite{14}, SPIRIT \cite{15}, SENSE \cite{70}, and MMF-Net \cite{6}. As shown in Table \ref{table9}, HyA-T achieves the top score in eight out of the 11 attributes, including DEF, FM, IPR, IV, OCC, OPR, OV, and SV. Specifically, for the IV, OCC, and OV, HyA-T demonstrates superior tracking accuracy, with improvements of 3.2$\%$, 3.2$\%$ and 3.6$\%$ on AUC, respectively. For the BC and MB attributes, HyA-T ranks the second, outperforming the other HS trackers. Overall, the experimental results demonstrate that HyA-T excels in handling various attributes and localizing the target with greater accuracy.

\subsection {Visualizations of tracking results}

To further illustrate the superior performance of HyA-T, Fig.~\ref{Fig6} presents the visualization results of HyA-T and other SOTA HS trackers, i.e., Trans-DAT \cite{71}, MMF-Net \cite{6}, SENSE \cite{70}, SEE-Net \cite{14}, PHTrack \cite{21}, and SPIRIT \cite{15}, on the HOTC2024 dataset \cite{24}. The scenarios in Fig.~\ref{Fig6} are selected from HOTC2024-NIR, HOTC2024-RedNIR, and HOTC2024-VIS data, specifically from the following videos: nir-rider10, nir-rider15, rednir-cloth3, rednir-dice3, vis-cup3, and vis-oranges5. As shown in Fig.~\ref{Fig6}, HyA-T tracks the target more accurately than the compared trackers. In nir-rider10 and nir-rider15, the other trackers fail to track the target due to small targets and the occlusion. And, in rednir-cloth3, rednir-dice3, vis-cup3, and vis-oranges5, the other trackers are interrupted by objects in the background, and inaccurately tracking the objects in the background as the targets. However, HyA-T tracks the targets accurately in all the sequences since the more effective spectral information utilization of the proposed methods. Overall, the qualitative results validate the effectiveness of the proposed methods in HyA-T, indicating its robustness and precision for HS tracking.

% \vspace{75 mm}
\section{Conclusion}
In this work, a new parameter-efficient fine-tuning method for the HS object tracking, HyA-T, is proposed to adapt the pretrained network to the task with undivided HS images. The proposed HAS and HAM modify the MSA and MLP in the pretrained network with adapters generated from the HS images, which reduces the limitation of performance improvement caused by the loss of the spectral information. Additionally, HEI further augments the spectral information into the input of the network to generates the input with undivided spectral information. Meanwhile, only the parameters in the proposed modules are fine-tuned, which is more efficient than the existing HS object tracking methods. Extensive experiments on four HS object tracking benchmarks with various spectral bands are conducted, and significant performance improvement is gained by the proposed methods. The HyA-T achieves state-of-the-art performance on all the four benchmarks, 0.705 on AUC and 0.958 on DP@20 in HOTC, 0.776 on AUC and 0.935 on DP@20 in HOTC2024-NIR.

\emph{Limitation.}
Despite the performance improvement in the HS object tracking task, HyA-T generates the adapters only with the HS images. However, the appearance change of the target, the scale information and the information in the frequency domain are not explored. Generating more effective spectral adapter from these information in an efficient manner may have potential to further improve the performance of the HS object tracking methods.

\section{Acknowledgments}
This work is supported by the project CEIEC-2022-ZM02-0247 and the 2023 Research Project of Shaanxi Provincial Department of Transportation, No.23-58X.

% or


% ============================================
%\appendices
%Appendix one text goes here %\cite{Roberg2010}.

%\section{}

%\section*{Acknowledgment}

%The authors would like to thank D. Root for the loan of the SWAP. The SWAP that can ONLY be usefull in Boulder...

\ifCLASSOPTIONcaptionsoff
  \newpage
\fi



%\begin{thebibliography}{1}


\bibliographystyle{IEEEtran}
\bibliography{HyA-T.bib}

\vspace{-10 mm}
\begin{IEEEbiography}[{\includegraphics[width=0.9in,clip,keepaspectratio]{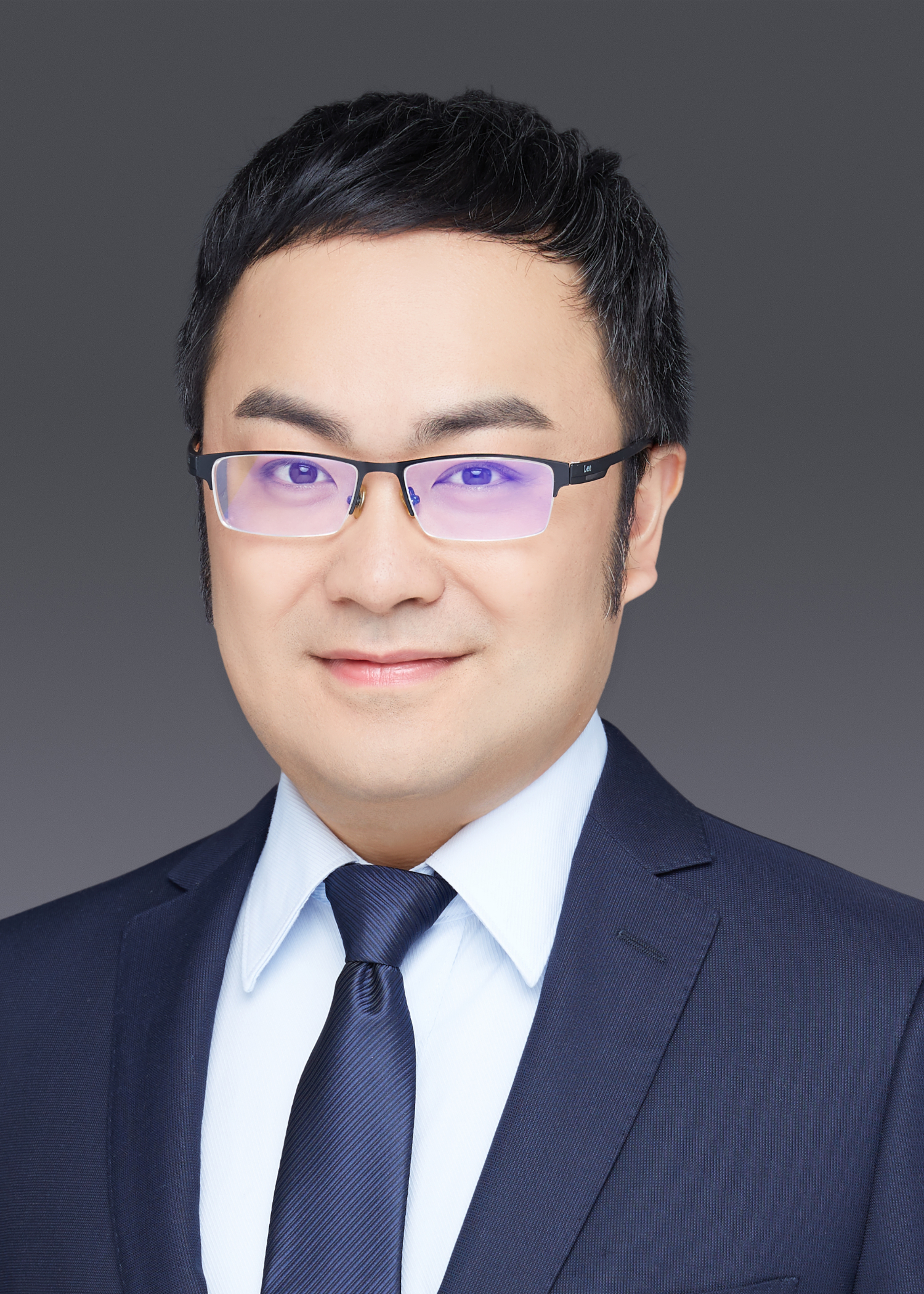}}]{Long Gao} (Member, IEEE) received the B.S. degree and the M.S. degree in Control Theory and Control Engineering from Xi'an Jiaotong University in 2010 and 2013. He received the Ph.D. degree in communication and information systems with Xidian University in 2019. Currently, he is a lecturer at communication and information systems with Xidian University. \par
His research interests include neural networks, deep learning and visual object tracking.
\end{IEEEbiography}

\vspace{-10 mm}
\begin{IEEEbiography}[{\includegraphics[width=0.9in,clip,keepaspectratio]{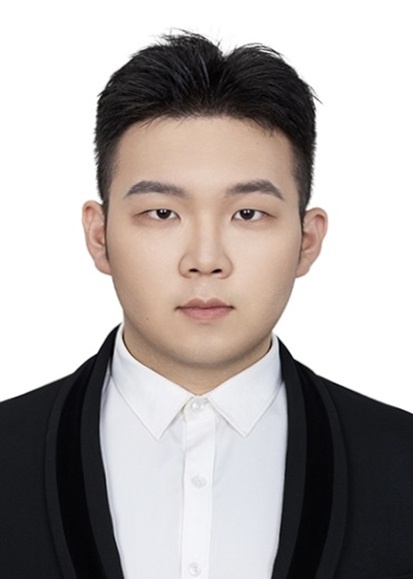}}]{Yunhe Zhang} received the B.S. degree in Communication engineering from the Tianjin University of Technology, Tianjin, China, in 2024. He is currently pursuing the M.S. degree with the Image Coding and Processing Center at State Key Laboratory of Integrated Services Networks, Xidian University, Xi’an, China. \par
His research interests include deep learning, neural network and hyperspectral object tracking.
\end{IEEEbiography}

\vspace{-10 mm}
\begin{IEEEbiography}[{\includegraphics[width=0.9in,clip,keepaspectratio]{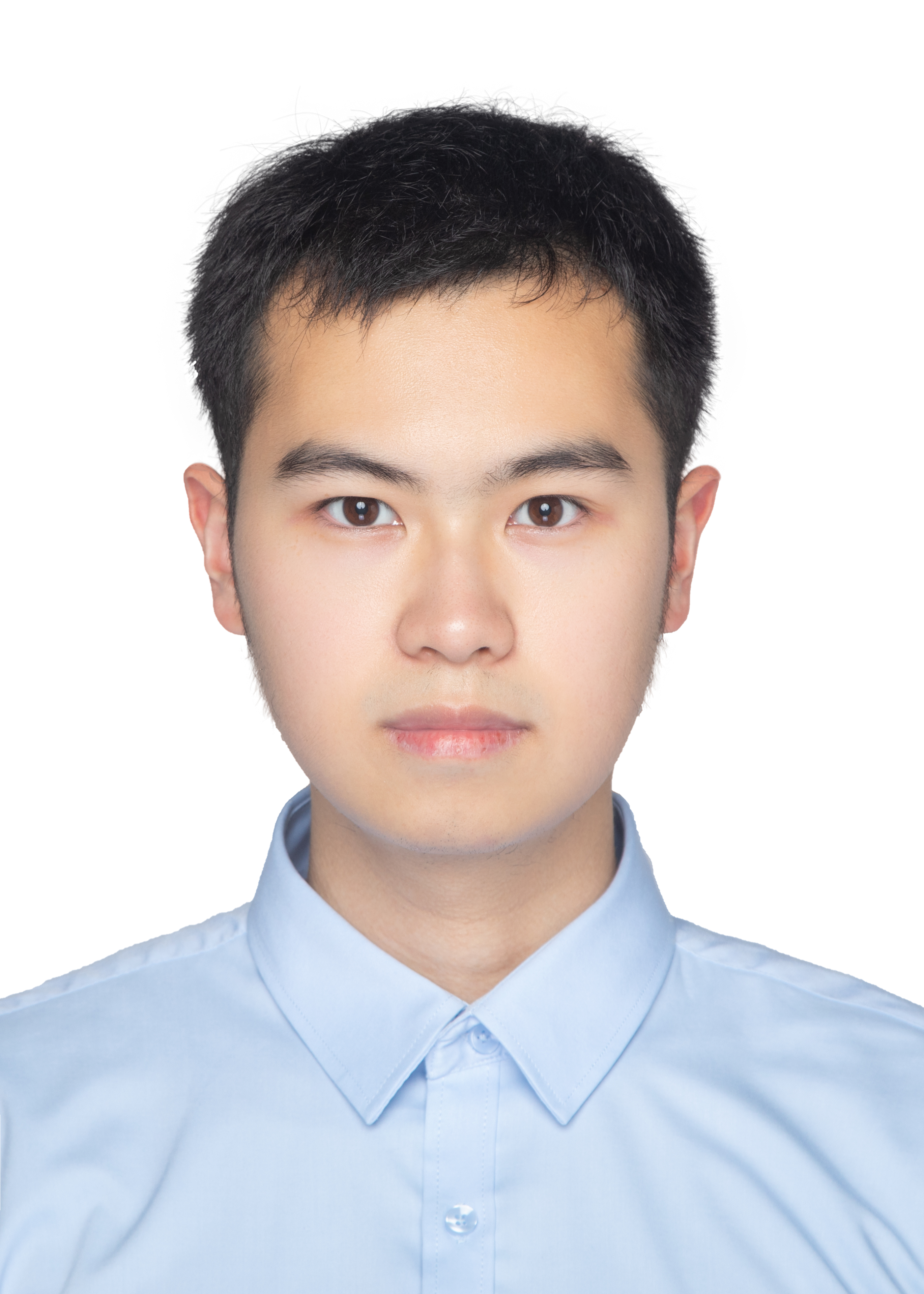}}]{Langkun Chen} received the B.E. degree in Communication engineering from the Ningbo University, Ningbo, China, in 2022. He is currently pursuing the M.S. degree with the Image Coding and Processing Center at State Key Laboratory of Integrated Services Networks, Xidian University, Xi’an, China. \par
His research interests include deep learning, neural network and hyperspectral object tracking.
\end{IEEEbiography}

\vspace{-10 mm}
\begin{IEEEbiography}[{\includegraphics[width=0.9in,clip,keepaspectratio]{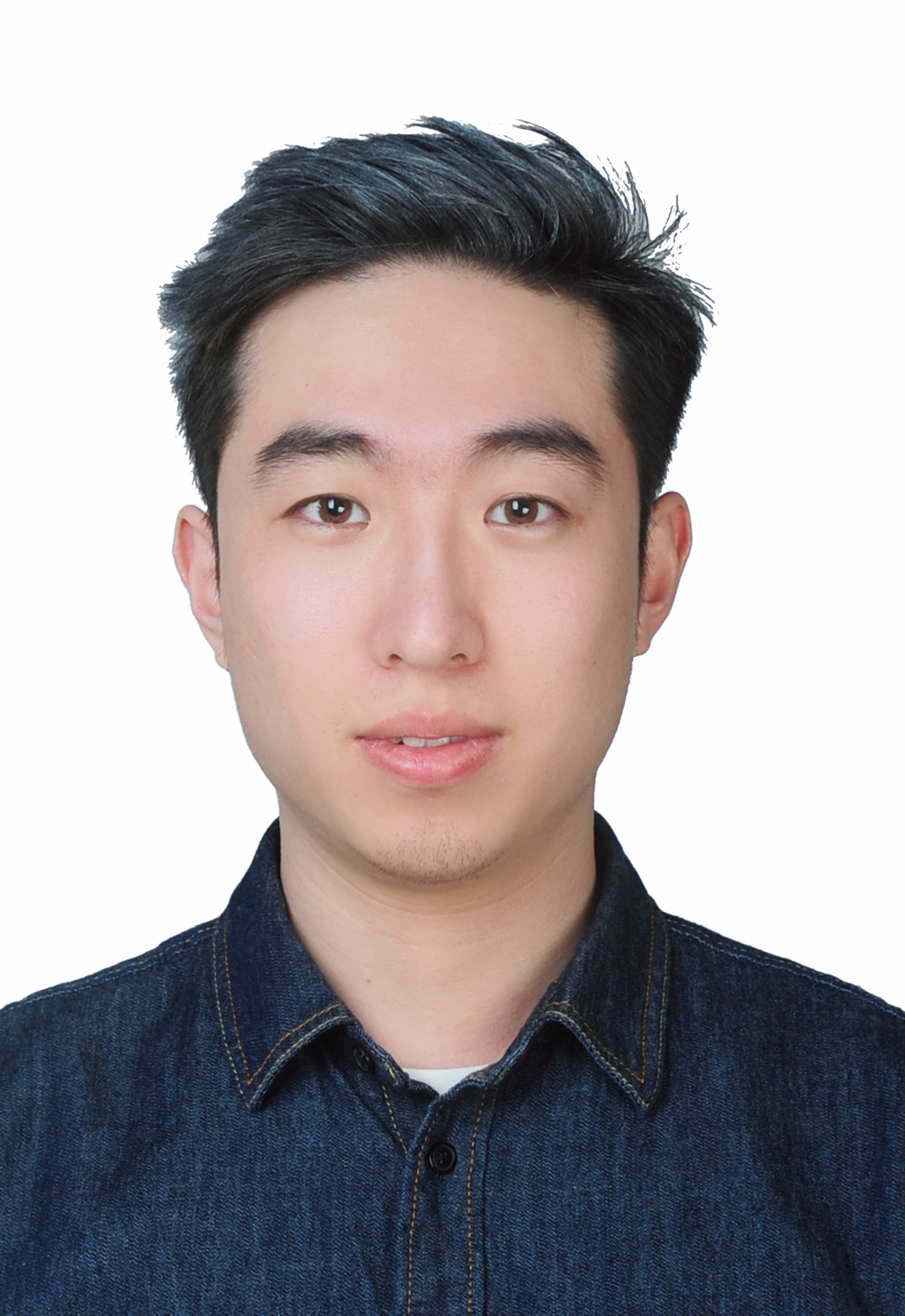}}]{Yan Jiang} received the B.E. degree in telecommunications engineering from the Xi’dian University, Xi’an, China, in 2016, and the M.Sc. degree in 2018 from the University of Sheffield, Sheffield, UK, where he is currently pursuing the Ph.D. degree with the Department of Electronic and Electrical Engineering. \par
His current research interests include wireless channel modelling, modulation system, mobile edge computing, smart environment modelling and deep learning.
\end{IEEEbiography}

\vspace{-10 mm}
\begin{IEEEbiography}[{\includegraphics[width=0.9in,clip,keepaspectratio]{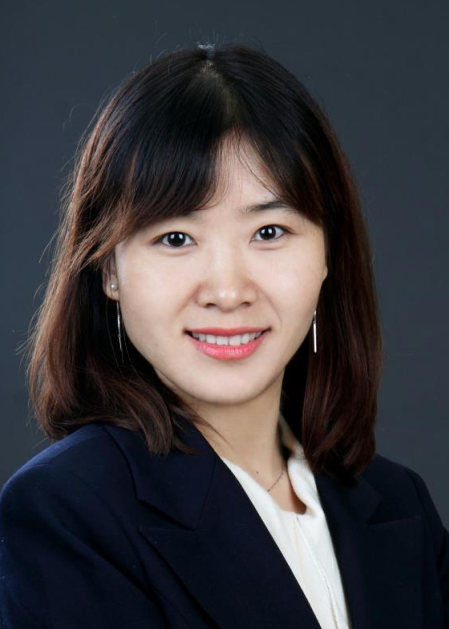}}]{Weiying Xie} (Senior Member, IEEE) received a B.S. degree in electronic information science and technology from the University of Jinan in 2011. She received an M.S. degree in communication and information systems, Lanzhou University in 2014, and the Ph.D. degree in communication and information systems of Xidian University in 2017. Currently, she is an Associate Professor with the State Key Laboratory of Integrated Services Networks, Xidian University. She has published more than 30 papers in refereed journals, including the IEEE TRANSACTIONS ON GEOSCIENCE AND REMOTE SENSING, the IEEE TRANSACTIONS ON NEURAL NETWORKS AND LEARNING SYSTEMS, the NEURAL NETWORKS, and the PATTERN RECOGNITION. \par
Her research interests include neural networks, machine learning, hyperspectral image processing, and high-performance computing.
\end{IEEEbiography}

\vspace{-10 mm}
\begin{IEEEbiography}[{\includegraphics[width=0.9in,clip,keepaspectratio]{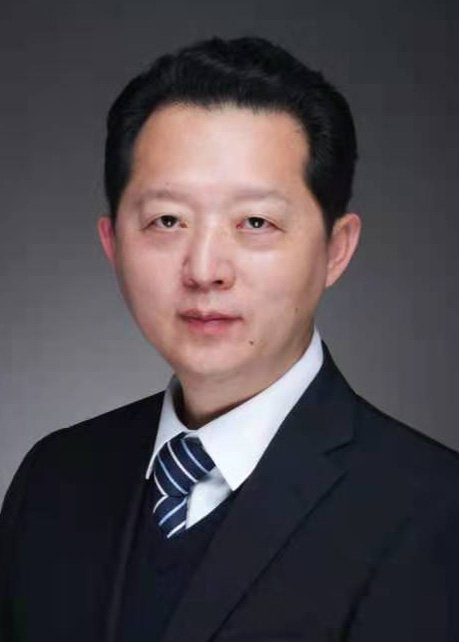}}]{Yunsong Li} (Member, IEEE) received the M.S. degree in telecommunication and information systems and the Ph.D. degree in signal and information processing from Xidian University, China, in 1999 and 2002, respectively. He joins the school of telecommunications Engineering, Xidian University in 1999 where he is currently a Professor. Prof. Li is the director of the image coding and processing center at the State Key Laboratory of Integrated Services Networks. \par
His research interests focus on image and video processing and high-performance computing.
\end{IEEEbiography}

\vfill


\end{document}